\documentclass[11pt]{article}
\PassOptionsToPackage{hyphens}{url}
\usepackage[preprint]{acl}

\usepackage{titlesec}
\titlespacing*{\subsection}{0pt}{5pt}{3pt}
\usepackage{times}
\usepackage{latexsym}
\usepackage[T1]{fontenc}
\usepackage[utf8]{inputenc}
\usepackage{microtype}
\usepackage{inconsolata}
\usepackage{graphicx}
\usepackage{xspace}
\usepackage{booktabs}
\usepackage{multirow}
\usepackage{algorithm}
\usepackage{algorithmic}
\graphicspath{{figures/}}

\usepackage{amsmath}
\usepackage{amssymb}

\newcommand{\method}{HRS\xspace}

\newcommand{\PHh}{\widehat{P}_H}

\title{Human-Residual Steering: Aligning Frozen LLM Judges
with Human Judgment Distributions}

\author{
  Chen Lyu \quad Phong Do \quad Gabriele Pergola \\
  Department of Computer Science \\
  University of Warwick, Coventry, UK \\
  \texttt{\{chen.lyu,Phong.Do,Gabriele.Pergola.1\}@warwick.ac.uk}
}

\begin{document}
\maketitle

\begin{abstract}
Large language models (LLMs) are widely used as automated judges, yet
most alignment methods target aggregated human judgments and overlook
variation among individual annotators. Existing approaches to matching
the full distribution of human ratings mainly rely on output-level calibration
or parameter fine-tuning. We propose \textbf{Human-Residual Steering
(HRS)}, to our knowledge the first method to align frozen LLM judges
with human judgment distributions through learned hidden-state
interventions. An oracle analysis shows that low-rank corrections at a
single layer remove most of the discrepancy between judge and human
distributions. HRS learns to predict such corrections from the judge's
hidden representation and rating logits, training on individual
annotations with proper scoring rules for ordinal and nominal judgments
and optimising through the frozen downstream layers. Across 14
benchmarks and four judges from three model families, HRS outperforms
all evaluated frozen-weight baselines. With Qwen2.5-7B, it achieves the
lowest Jensen-Shannon distance on 13 of 14 benchmarks, reducing the
mean from 0.565 to 0.352 versus 0.377 for the strongest calibrator. HRS
also raises mean Spearman correlation with aggregate human scores from
0.449 to 0.579, improves population-level estimation under limited
annotation budgets, and trains about 88 times fewer parameters than
distributional fine-tuning at a modest cost in alignment. \footnote{Code will be released upon acceptance.}

\end{abstract}

\section{Introduction}
\label{sec:introduction}

Large language models (LLMs) are increasingly employed as automated
judges of response quality, helpfulness, safety, and relevance,
providing a scalable alternative to repeated human evaluation
\citep{zheng2023judging,geval}.
However, existing judge-alignment methods predominantly assess
agreement with aggregated human judgments, such as mean ratings
or majority labels, using metrics including rank correlation,
prediction error, and classification accuracy
\citep{geval,lager,saja}.
While these measures capture agreement with human consensus,
they do not establish whether a judge reflects the distribution
of individual human judgments \citep{baan2022,plank2022}.

This distinction is particularly important in subjective
evaluation, where annotators may differ in their interpretations,
preferences, and perspectives \citep{uma2021,plank2022,baan2022}.
Such variation arises in both \emph{ordinal} judgments, whose
categories have a meaningful ordering, and \emph{nominal}
judgments, whose categories are unordered.
For example, SummEval reports substantial variation
among expert annotators (Krippendorff's $\alpha=0.413$
in the initial annotation round) \citep{summeval},
while ChaosNLI finds that original majority labels
differ from those obtained with 100 annotations
per example for 25\% of SNLI and 32\% of MNLI
examples in its selected subsets \citep{chaosnli}.
Despite growing recognition of human disagreement, aligning
LLM judges with the full distribution of individual human
ratings remains comparatively underdeveloped.
Consequently, judges may achieve strong agreement with
aggregated human judgments while failing to capture
the extent and structure of annotator disagreement.

Recent efforts to address this gap have begun to target
\emph{human judgment distributions} (HJDs) directly.
Distributional LLM-as-a-Judge adapts model parameters to
match empirical human ratings \citep{chen2025distributional},
while NAPHA calibrates judge outputs towards human soft
labels \citep{napha}.
However, these approaches primarily rely on parameter
adaptation or post-hoc output transformations, leaving
representation-level distributional alignment largely
unexplored.
In particular, whether human-judge distributional
discrepancies can be corrected within a frozen judge's
hidden representations remains an open question.

In parallel, representation-level methods have demonstrated
that LLM behaviour can be modified through hidden-state
interventions without updating the backbone \citep{caa,reft}.
For LLM judges, intermediate representations have been shown
to encode evaluation-related information beyond final predictions
\citep{lager}, while low-dimensional directions have been
associated with judge biases \citep{unfairjudge}.
These findings suggest that internal representations may offer
an alternative to output-level calibration for human-distribution
alignment. However, it remains unclear whether compact
hidden-state corrections can recover human judgment distributions
and, crucially, whether such corrections can generalise to
unseen items without access to their human ratings.

To investigate these questions, we propose \textbf{Human-Residual
Steering (HRS)}, to the best of our knowledge, the first approach
to align frozen LLM judges with human judgment distributions
through learned hidden-state interventions.
We begin by deriving \emph{oracle human-judge residuals},
which use observed human ratings to identify hidden-state
corrections that reduce distributional discrepancies.
Our analysis reveals that these corrections exhibit a
predominantly low-rank structure, with compact subspaces
preserving most of the oracle alignment gain.
This finding motivates HRS, which learns to predict
input-conditioned corrections within a low-dimensional
intervention space, without requiring human ratings at inference.

Specifically, HRS learns and predicts corrections from the judge's
hidden representations and original rating logits, applying
them at a single transformer layer.
The interventions are trained using individual human
annotations and proper scoring rules for ordinal and
nominal judgments, with gradients propagated through
the judge's frozen downstream layers. HRS optimises
the resulting judgment distributions while keeping
all backbone parameters fixed.

We evaluate HRS across 14 criterion-level benchmarks from 8 dataset families, spanning ordinal and nominal judgments and 4 LLM judges. HRS consistently outperforms evaluated output-level calibration methods, achieving the lowest Jensen–Shannon distance on 13 of 14 benchmarks with Qwen2.5-7B and reducing the mean from 0.565 to 0.352, compared with 0.377 for the strongest output-level calibrator. These improvements generalise across judge families. Beyond distributional alignment, HRS achieves the highest aggregate-score correlation among evaluated frozen-backbone methods and improves the precision of downstream population-level estimation with limited evaluation-stage human annotations. Further comparisons reveal an accuracy–efficiency trade-off with distributional fine-tuning, which achieves stronger alignment at the cost of updating substantially more parameters.

Our main contributions are threefold:
\begin{enumerate}
    \item \textbf{Representation-level distributional alignment.}
    We provide the first systematic investigation of
    hidden-state interventions for aligning frozen LLM judges
    with human judgment distributions, demonstrating through
    oracle analysis that low-rank corrections can
    substantially reduce distributional discrepancies.

    \item \textbf{Human-Residual Steering (HRS).}
    We introduce an input-conditioned, low-rank intervention
    that learns from individual human ratings using
    ordinal- and nominal-aware objectives, optimised through
    the judge's frozen downstream layers.

    \item \textbf{Extensive empirical evaluation.}
    Across 14 ordinal and nominal benchmarks and multiple judge families, HRS outperforms evaluated output-level calibration methods, achieves strong aggregate-score agreement among frozen-weight methods, and improves downstream statistical estimation, while revealing accuracy–efficiency trade-offs with parameter fine-tuning.
\end{enumerate}

\begin{figure*}[t]
    \centering
    \includegraphics[width=0.9\textwidth]{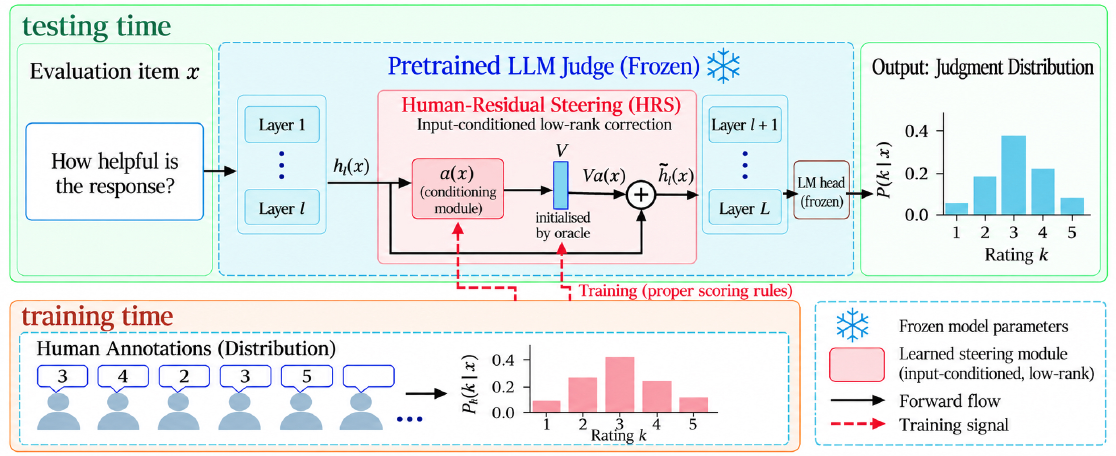}
    \caption{
        \textbf{Overview of Human-Residual Steering (HRS).}
        Given an evaluation item $x$, HRS predicts an
        input-conditioned, low-rank correction
        $\alpha V a(x)$ from the frozen judge's hidden
        representation and original rating logits.
        The correction is applied at a single transformer
        layer, and the modified representation is propagated
        through the remaining frozen layers to produce
        an aligned human judgment distribution.
    }
    \label{fig:hrs_overview}
\end{figure*}

\section{Related Work}
\label{sec:literature}

\paragraph{Aggregate-Level LLM Judge Alignment.}
Existing approaches predominantly align LLM judges with
aggregated human evaluations, such as mean ratings or
consensus labels.
These include rubric- and reasoning-based prompting
\citep{geval,hypoeval}, post-hoc output calibration
\citep{vandenburg2025,saja}, and representation-based
judge alignment \citep{lager}.
While effective at improving agreement with aggregated
human judgments, these methods do not explicitly target
the distribution of individual human ratings.

\paragraph{Distributional LLM Judge Alignment.}
Recent work has begun to align LLM judges with
\emph{human judgment distributions} (HJDs).
\citet{chen2024hjd} investigate approximating HJDs
using limited expert annotations and explanations,
while \citet{chen2025distributional} fine-tune LLM judges
to match empirical human rating distributions.
More recently, \citet{napha} propose entropy-aware
post-hoc calibration towards human soft labels.
This research builds on the broader literature on
\emph{human label variation} (HLV), which emphasises
preserving annotator disagreement rather than collapsing
individual annotations into a single ground truth
\citep{uma2021,plank2022,baan2022,meissner2021}.
Unlike existing distributional alignment approaches,
HRS learns input-conditioned corrections within a frozen
judge's hidden representations.

\paragraph{Representation-Level Adaptation and Steering.}
Prior work has shown that hidden-state interventions can
modify LLM behaviour without updating backbone parameters
\citep{iti,caa}.
Representation Fine-Tuning (ReFT) further enables learned,
parameter-efficient interventions \citep{reft}.
For LLM judges, LAGER \citep{lager} leverages intermediate
representations to improve human-judge agreement, while
\citet{unfairjudge} identify low-dimensional subspaces
associated with judge biases.
However, their potential for aligning judges with human
judgment distributions remains underexplored.

\section{Methodology}
\label{sec:method}

We introduce Human-Residual Steering (HRS), which aligns a frozen
LLM judge with human rating distributions through input-conditioned
hidden-state interventions. We first formalise distributional
judgment prediction (\S\ref{sec:setting}), then examine oracle
corrections to characterise representation-level correctability
(\S\ref{sec:oracle}). Building on this analysis, we describe HRS
(\S\ref{sec:hrs}) and its distributional training objective
(\S\ref{sec:training}).

\subsection{Distributional Human Judgment}
\label{sec:setting}

\paragraph{Human judgment distribution.}
We consider evaluation tasks where multiple annotators assign
potentially different ratings to the same item~\citep{plank2022}.
Let $x$ denote an evaluation item and its criterion, with
$m_x$ annotations $y_{x,1},\ldots,y_{x,m_x}\in\mathcal{Y}$,
where $\mathcal{Y}=\{0,\ldots,K-1\}$ represents either ordinal
ratings or nominal categories. The empirical human judgment
distribution is
\begin{equation}
P_H(k\mid x)
=
\frac{1}{m_x}
\sum_{j=1}^{m_x}
\mathbb{1}[y_{x,j}=k],
\qquad k\in\mathcal{Y}.
\label{eq:ph}
\end{equation}
As this distribution is estimated from finitely many annotations,
it may differ from the underlying population distribution.

\paragraph{Frozen LLM judge.}
Let $J$ be a decoder-only language model with $L$ transformer
blocks and hidden dimension $d$. Given a fixed evaluation prompt
and $K$ rating options, each represented by a single-token
verbalizer $t_k$, we extract the corresponding next-token logits
at the final prompt position:
\[
z(x)
=
\big(
u_{t_0}(x),\ldots,u_{t_{K-1}}(x)
\big)
\in\mathbb{R}^{K}.
\]
The judge's rating distribution is obtained by normalising over
the valid rating tokens:
\begin{equation}
P_J(k\mid x)
=
\frac{\exp z_k(x)}
     {\sum_{k'=0}^{K-1}\exp z_{k'}(x)}.
\label{eq:pj}
\end{equation}
This formulation does not require free-form generation.

Let $h_l(x)\in\mathbb{R}^{d}$ denote the residual-stream
representation at the final prompt position after block $l$.
We write $J_{l+1:L}(h;x)$ for the rating distribution produced
by replacing $h_l(x)$ with $h$ and propagating it through the
remaining frozen blocks. Accordingly,
\[
J_{l+1:L}(h_l(x);x)=P_J(\cdot\mid x).
\]

\subsection{Oracle Human-Judge Residuals}
\label{sec:oracle}

We first investigate whether the discrepancy between a frozen
judge and an empirical human rating distribution can be
corrected through a single hidden-state intervention.
For an item $x$, we define the oracle human-judge residual
$\delta^\ast(x)\in\mathbb{R}^{d}$ as a minimum-norm
correction that approximately matches the target distribution:
\begin{equation}
\begin{split}
\delta^\ast(x)
\in \mathop{\mathrm{arg\,min}}\limits_{\delta}
&\ \|\delta\|_2, \\
\text{s.t.}\qquad
J_{l+1:L}(h_l(x)+\delta;x)
&\approx \widetilde P_H(\cdot\mid x),
\end{split}
\label{eq:oracle}
\end{equation}
where $\widetilde P_H$ is a smoothed empirical distribution
used for numerical stability.

Because the oracle uses the same item's human ratings,
it serves as a diagnostic of representation-level
correctability rather than a deployable predictor.
We approximate Eq.~\ref{eq:oracle} using a local
linearisation of the frozen downstream network followed
by a damped iterative logit-correction procedure
(Appendix~\ref{app:oracle}).

\paragraph{Low-rank residual subspace.}
To examine whether the oracle corrections share a compact
structure, we apply singular value decomposition (SVD)
to the matrix of training-item residuals. The leading
$r$ right singular vectors define an initial intervention
basis,
\begin{equation}
V^{(0)}=[w_1,\ldots,w_r]\in\mathbb{R}^{d\times r}.
\label{eq:oracle_basis}
\end{equation}
Concentration of residual energy in the leading directions
indicates a shared low-dimensional correction subspace.
Only training-item residuals are used to initialise this
basis.

\subsection{Human-Residual Steering}
\label{sec:hrs}

Oracle corrections require item-level human ratings and therefore cannot be used directly at inference time. HRS instead learns an input-conditioned intervention within a low-rank subspace, as illustrated in Figure~\ref{fig:hrs_overview}:

\begin{equation}
h'_l(x)
=
h_l(x)+\alpha Va(x),
\label{eq:intervention}
\end{equation}
where $V\in\mathbb{R}^{d\times r}$ is a trainable
intervention basis, $a(x)\in\mathbb{R}^{r}$ contains
item-specific coefficients, and $\alpha$ controls the
intervention scale. The modified representation is passed
through the judge's frozen downstream computation to produce
\begin{equation}
\widehat P_H(\cdot\mid x)
=
J_{l+1:L}
\big(h_l(x)+\alpha Va(x);x\big).
\label{eq:hrspred}
\end{equation}
This follows the representation-intervention setting
of~\citet{reft}, while targeting the distribution of
individual human judgments.

We initialise $V$ from the oracle basis $V^{(0)}$ and allow
it to remain trainable. The coefficients are predicted by
a lightweight linear router:
\begin{align}
a(x) &= W_a\phi(x)+b_a,
\label{eq:router}\\
\phi(x) &=
\left[
\operatorname{LN}(h_l(x));
\widetilde z(x)
\right],
\end{align}
where $\operatorname{LN}$ is parameter-free normalisation
and $\widetilde z(x)$ denotes the centred and standardised
rating logits of the original judge. The intervention is
therefore conditioned on both the internal representation
and the judge's initial rating tendency.

HRS is optimised through the original frozen downstream
blocks $J_{l+1:L}$, with gradients updating only
$V$, $W_a$, and $b_a$. Training uses $\alpha=1$; the scale
$\alpha$ is selected afterwards on the validation split
(Appendix~\ref{app:training}). Importantly, the oracle residuals
are used for initialisation rather than as regression
targets: HRS learns corrections according to their effect
on the final judgment distribution.

\subsection{Learning from Individual Human Ratings}
\label{sec:training}

We train HRS using individual human annotations rather than
aggregated consensus labels. For a mini-batch $\mathcal{B}$,
the objective is
\begin{align}
\bar S(x)
&=
\frac{1}{m_x}
\sum_{j=1}^{m_x}
S\!\left(
\widehat P_H(\cdot\mid x),y_{x,j}
\right),
\label{eq:itemscore}\\
\mathcal{L}
&=
\frac{1}{|\mathcal{B}|}
\sum_{x\in\mathcal{B}}
\left[
\bar S(x)+\lambda_a\|a(x)\|_2^2
\right],
\label{eq:loss}
\end{align}
where $S(P,y)$ is a proper scoring rule and the second
term regularises the intervention coefficients. Averaging
annotations within each item ensures equal weighting
regardless of the number of raters.

We use the ranked probability score (RPS) for ordinal
judgments and cross-entropy (CE) for nominal judgments.
RPS incorporates the ordering of rating categories
through their cumulative probabilities, whereas CE
does not assume an ordering:
\begin{align}
S_{\mathrm{RPS}}(P,y)
&=
\sum_{k=0}^{K-2}
\left(
F_P(k)-\mathbb{1}[y\leq k]
\right)^2,
\label{eq:rps}\\
S_{\mathrm{CE}}(P,\widetilde P_H)
&=-\sum_{k=0}^{K-1}\widetilde P_H(k)\log P(k),
\label{eq:ce}
\end{align}
where $F_P(k)=\sum_{k'=0}^{k}P(k')$ denotes the cumulative
probability up to category $k$. For nominal judgments, the
per-annotation average in Eq.~\eqref{eq:itemscore} is replaced by
$S_{\mathrm{CE}}$ evaluated against the smoothed histogram
$\widetilde P_H(\cdot\mid x)$. The smoothing shrinks the
empirical histogram towards the training marginal, with its
strength selected by validation leave-one-out RPS
(Appendix~\ref{app:training}). Both objectives provide distributional supervision from individual ratings.
Alternative objectives are examined in
\S\ref{sec:ablation} and Appendix~\ref{app:scores}.

\paragraph{Training and inference.}
The judge's parameters, prompt, verbalizers, and output
head remain fixed throughout training. We cache the
unchanged prefix computation and optimise the intervention
through the remaining frozen layers. At inference, HRS
first obtains the judge's hidden representation and original
rating logits, then applies the predicted correction and
re-evaluates the frozen downstream layers. This introduces
an additional suffix computation but requires neither human
ratings nor test-time optimisation. Implementation details
are provided in Appendices~\ref{app:oracle}, \ref{app:training},
and~\ref{app:details}.


\begin{table*}[t]
\centering
\scriptsize
\setlength{\tabcolsep}{3.2pt}
\begin{tabular}{@{}l*{14}{c}c@{}}
\toprule
\multirow{2}{*}{Method}
& \multicolumn{4}{c}{\scriptsize\bfseries SummEval}
& \multicolumn{4}{c}{\scriptsize\bfseries Topical-Chat}
& \multirow{2}{*}{FLASK}
& \multirow{2}{*}{HS2}
& \multirow{2}{*}{HS3}
& \multirow{2}{*}{CNLI}
& \multirow{2}{*}{DynaS}
& \multirow{2}{*}{Scrup}
& \multirow{2}{*}{Avg.\ $\downarrow$} \\
[1pt]
& Coh & Con & Flu & Rel
& Ovr & Nat & Ctx & Eng
& & & & & & & \\
[0pt]
\midrule
Raw judge
& .609$^{\dagger}$ & .772$^{\dagger}$ & .786$^{\dagger}$ & .590$^{\dagger}$
& .594$^{\dagger}$ & .492$^{\dagger}$ & .500$^{\dagger}$ & .411$^{\dagger}$
& .572$^{\dagger}$ & .542$^{\dagger}$ & .641$^{\dagger}$
& \multicolumn{1}{|c}{.371$^{\dagger}$}
& .342$^{\dagger}$ & .683$^{\dagger}$
& \multicolumn{1}{|c}{.565} \\

TS-Soft
& .512$^{\dagger}$ & .622$^{\dagger}$ & .615$^{\dagger}$ & .500$^{\dagger}$
& .499$^{\dagger}$ & .426$^{\dagger}$ & .419$^{\dagger}$ & .385$^{\dagger}$
& .511$^{\dagger}$ & .469$^{\dagger}$ & .608$^{\dagger}$
& \multicolumn{1}{|c}{.220$^{\dagger}$}
& .360$^{\dagger}$ & .517$^{\dagger}$
& \multicolumn{1}{|c}{.476} \\

PTS
& .512$^{\dagger}$ & .614$^{\dagger}$ & .615$^{\dagger}$ & .499$^{\dagger}$
& .496$^{\dagger}$ & .420$^{\dagger}$ & .408$^{\dagger}$ & .384$^{\dagger}$
& .507$^{\dagger}$ & .461$^{\dagger}$ & .608$^{\dagger}$
& \multicolumn{1}{|c}{.220$^{\dagger}$}
& .350$^{\dagger}$ & .513$^{\dagger}$
& \multicolumn{1}{|c}{.472} \\

Dirichlet (ODIR)
& .473$^{\dagger}$ & .209$^{\dagger}$ & .288$^{\dagger}$ & .401$^{\dagger}$
& .473$^{\dagger}$ & \textbf{.376} & .352 & .345$^{\dagger}$
& .425$^{\dagger}$ & .448$^{\dagger}$ & .576$^{\dagger}$
& \multicolumn{1}{|c}{.214$^{\dagger}$}
& .324$^{\dagger}$ & .375$^{\dagger}$
& \multicolumn{1}{|c}{.377} \\

Linear mapper
& .477$^{\dagger}$ & .214$^{\dagger}$ & .294$^{\dagger}$ & .404$^{\dagger}$
& .482$^{\dagger}$ & .390 & .369$^{\dagger}$ & .374$^{\dagger}$
& .435$^{\dagger}$ & .453$^{\dagger}$ & .601$^{\dagger}$
& \multicolumn{1}{|c}{.253$^{\dagger}$}
& .373$^{\dagger}$ & .380$^{\dagger}$
& \multicolumn{1}{|c}{.393} \\

MLP mapper
& .485$^{\dagger}$ & .332$^{\dagger}$ & .365$^{\dagger}$ & .424$^{\dagger}$
& .512$^{\dagger}$ & .411$^{\dagger}$ & .397$^{\dagger}$ & .404$^{\dagger}$
& .444$^{\dagger}$ & .456$^{\dagger}$ & .601$^{\dagger}$
& \multicolumn{1}{|c}{.243$^{\dagger}$}
& .371$^{\dagger}$ & .383$^{\dagger}$
& \multicolumn{1}{|c}{.416} \\

NAPHA
& .498$^{\dagger}$ & .539$^{\dagger}$ & .529$^{\dagger}$ & .466$^{\dagger}$
& .513$^{\dagger}$ & .413$^{\dagger}$ & .397$^{\dagger}$ & .407$^{\dagger}$
& .498$^{\dagger}$ & .455$^{\dagger}$ & .600$^{\dagger}$
& \multicolumn{1}{|c}{.227$^{\dagger}$}
& .354$^{\dagger}$ & .383$^{\dagger}$
& \multicolumn{1}{|c}{.449} \\

NAPHA (log-prob.)
& .483$^{\dagger}$ & .270$^{\dagger}$ & .314$^{\dagger}$ & .413$^{\dagger}$
& .489$^{\dagger}$ & .408$^{\dagger}$ & .382$^{\dagger}$ & .382$^{\dagger}$
& .445$^{\dagger}$ & .450$^{\dagger}$ & .582$^{\dagger}$
& \multicolumn{1}{|c}{.221$^{\dagger}$}
& .330$^{\dagger}$ & .379$^{\dagger}$
& \multicolumn{1}{|c}{.396} \\

\midrule
\textbf{HRS (ours)}
& \textbf{.414} & \textbf{.159} & \textbf{.241} & \textbf{.383}
& \textbf{.440} & .379 & \textbf{.349} & \textbf{.321}
& \textbf{.410} & \textbf{.432} & \textbf{.546}
& \multicolumn{1}{|c}{\textbf{.198}}
& \textbf{.302} & \textbf{.348}
& \multicolumn{1}{|c}{\textbf{.352}} \\
\bottomrule
\end{tabular}

\caption{
Human-distribution alignment across 14 criterion-level benchmarks using Qwen2.5-7B (test JSD averaged over three seeds; lower is better), compared with frozen-weight baselines. Avg. denotes the unweighted macro-average across criteria. Bold indicates the best result; † denotes a significant paired difference from HRS under prompt-level bootstrap resampling. Weight-updating fine-tuning is separately compared in \S\ref{sec:tradeoffs} and Appendix~\ref{app:finetune}.}
\label{tab:h1}
\end{table*}

\section{Experimental Setup}
\label{sec:experiments}
\label{sec:setup}

\paragraph{Benchmarks.}
We evaluate HRS on 14 criterion-level benchmarks from eight dataset
families (Appendix Table~\ref{tab:benchmarks}). Eleven involve ordinal
ratings: four SummEval dimensions~\citep{summeval}, four Topical-Chat
criteria~\citep{usr,topicalchat}, FLASK~\citep{flask,prometheus},
HelpSteer2 helpfulness~\citep{helpsteer2}, and HelpSteer3 code
preferences~\citep{helpsteer3}. The three nominal benchmarks are
ChaosNLI~\citep{chaosnli}, DynaSent Round~2~\citep{dynasent}, and
Scruples Anecdotes~\citep{scruples}. All benchmarks provide multiple
individual human annotations per item. We use official splits where
available; otherwise, we construct 60/20/20 train/validation/test
splits, grouping related examples to prevent leakage. Dataset
statistics and splitting procedures are detailed in
Appendix~\ref{app:benchmarks}.

\paragraph{Judges.}
We use Qwen2.5-7B-Instruct~\citep{qwen25} as the primary judge,
evaluated over three experimental seeds (with split resampling
where applicable), and replicate the main
comparisons with Llama-3.1-8B-Instruct~\citep{llama3},
Gemma-2-9B-it~\citep{gemma2}, and Qwen2.5-14B-Instruct.
All judges use a fixed zero-shot evaluation prompt
(Appendix~\ref{app:prompt_template}). Model-specific coverage
is reported in Appendix Table~\ref{tab:coverage}.

\paragraph{Baselines.}
We organise comparisons by adaptation mechanism and evaluation objective.
For human-distribution alignment, our primary baselines are
\textbf{frozen-weight output-level calibration methods}, which preserve the judge
backbone and transform its predicted scores: temperature scaling
(TS-Soft) \citep{dogah2026}, parameterised temperature scaling
(PTS) \citep{pts}, Dirichlet calibration \citep{kull2019},
linear and MLP output mappers, and NAPHA \citep{napha}.
We additionally compare with our replication of the
\textbf{weight-updating distributional fine-tuning} method of
\citet{chen2025distributional}, implemented using LoRA
\citep{lora}. This provides a complementary comparison with
backbone adaptation rather than a frozen-backbone calibrator.
For conventional aggregate-score agreement, we compare with
G-Eval \citep{geval}, LAGER \citep{lager},
SAJA \citep{saja}, and HypoEval \citep{hypoeval}.
All methods use matched data splits and base judges where
applicable; implementation details are provided in
Appendices \ref{app:expbaselines} and \ref{app:finetune}.

\paragraph{Evaluation Metrics.}
Our primary metric is Jensen-Shannon distance (JSD),
defined as the square root of Jensen-Shannon divergence
between predicted and empirical human rating distributions,
using natural logarithms \citep{lin1991,endres2003}.
We additionally report annotation-level ranked probability
score (RPS) for ordinal ratings \citep{epstein1969}
and Brier score for nominal labels \citep{gneiting2007}.
Aggregate-score agreement is evaluated using Spearman's
$\rho$ and mean absolute error (MAE) for ordinal ratings,
and accuracy and macro-F1 for nominal labels.
Additional distributional and calibration metrics are
reported in Appendices~\ref{app:scores} and~\ref{app:expfull}.
Statistical significance is assessed using paired cluster
bootstrap confidence intervals (95\%), resampling at the
prompt or source-group level \citep{efron1993}.

\paragraph{Experimental Protocol.}
All methods are trained on the training split, with
hyperparameters selected using validation data or
training-set cross-validation. HRS uses annotation-level RPS for ordinal
judgments and cross-entropy to the smoothed histogram for nominal judgments. Main results are
averaged over three Qwen2.5-7B seeds, with other judges
evaluated using the reported configurations. In the cross-judge
comparison (Table~\ref{tab:judgesbench}), the best post-hoc
baseline is selected retrospectively by test JSD for each
benchmark, providing an optimistic baseline comparison.
Further implementation details and hyperparameters are
provided in Appendix~\ref{app:details}.

\section{Experimental Results}
\label{sec:results}

We first examine the feasibility of representation-level
alignment through oracle corrections (\S\ref{sec:oracleresults}),
then evaluate HRS against frozen-weight output-level calibration methods
across benchmarks and judge families (\S\ref{sec:main}).
We then assess its generalisation to aggregate-score evaluation
(\S\ref{sec:agg}) and ablate on loss function design choices (\S\ref{sec:ablation}). Finally, \S\ref{sec:tradeoffs} examines accuracy-efficiency
trade-offs against weight-updating fine-tuning and the downstream statistical utility of HRS.


\subsection{Representation-Level Correctability}
\label{sec:oracleresults}

We first investigate whether discrepancies between a frozen
LLM judge and human rating distributions can be corrected
through hidden-state interventions, and whether these
corrections exhibit a low-dimensional structure.
To assess the achievable correction, we derive oracle
residuals using each test item's human ratings
(\S\ref{sec:oracle}) and propagate the modified
representations through the judge's frozen downstream layers.
As these interventions use test-item annotations, they
measure representation-level correctability rather than
predictive generalisation.

\paragraph{A single-layer intervention substantially reduces
distributional discrepancies.}
Across 14 benchmarks, oracle corrections reduce mean JSD
from $0.565$ to $0.179$ for Qwen (mean of three seeds)
and from $0.490$ to $0.184$ for Llama (seed 42),
corresponding to relative reductions of 68\% and 63\%,
respectively.
These results demonstrate that substantial human-judge
distributional discrepancies can be corrected by modifying
representations at a single layer, without updating the
judge's parameters.
Moreover, interventions at depths beyond $0.6L$ achieve
comparable alignment (Appendix
Table~\ref{tab:oracledepthcomplete}), suggesting that
correctability is not restricted to a narrowly selected
intervention layer.

\paragraph{Low-dimensional subspaces retain most of the
corrective capacity.}
We next examine whether the oracle corrections share a
compact structure across items, motivating the low-rank
parameterisation of HRS.
Using SVD on training-item oracle residuals, we construct
an eight-dimensional subspace and project test-item oracle
corrections onto it.
Despite this dimensionality reduction, the leading eight
directions retain a median of 94\% of residual energy
for Qwen and at least 98.8\% for Llama
(Appendix Table~\ref{tab:oraclespectrumcomplete}).
The projected corrections achieve mean JSDs of $0.241$
and $0.185$, preserving approximately 84\% and 99\%
of the full oracle improvement, respectively.
This indicates that most of the achievable alignment gain
is concentrated in a small shared intervention subspace,
providing empirical support for HRS's low-rank design.
Importantly, these projected corrections still use
test-item ratings to determine intervention coefficients;
HRS subsequently addresses the challenge of predicting
such corrections without access to these ratings.

\begin{table}[t]
\centering\scriptsize\setlength{\tabcolsep}{3pt}
\resizebox{\columnwidth}{!}{%
\begin{tabular}{lccccc}
\hline
Judge & $n$ & Raw & Best post-hoc & HRS & Wins \\
\hline
Qwen2.5-7B & 14 & .565 & .377 & \textbf{.352} & 13/14 \\
Llama-3.1-8B & 14 & .490 & .387 & \textbf{.367} & 12/14 \\
Gemma-2-9B & 14 & .555 & .367 & \textbf{.350} & 12/14 \\
Qwen2.5-14B & 10 & .551 & .356 & \textbf{.336} & 9/10 \\
\hline
\end{tabular}}
\caption{Cross-judge summary: mean test JSD over $n$ evaluated benchmarks and HRS wins against the best post-hoc baseline. Best post-hoc is an optimistic envelope selected separately by test JSD for each benchmark. Complete per-benchmark numbers appear in Appendix Table~\ref{tab:judgesbenchfull}.}
\label{tab:judgesbench}
\end{table}

\subsection{Human-Distribution Alignment}
\label{sec:main}

\paragraph{HRS consistently outperforms frozen-weight baselines.}
Table~\ref{tab:h1} compares HRS with frozen-weight methods across 14 benchmarks using Qwen2.5-7B. HRS achieves the lowest mean JSD ($0.352$), substantially improving over the raw judge ($0.565$) and the strongest calibration baseline ($0.377$). It obtains the best alignment on 13 of 14 benchmarks, with statistically significant gains on 12. The only exception is Topical-Chat naturalness, where the difference is not significant. These results demonstrate the effectiveness of hidden-state interventions over output-level calibration for both ordinal and nominal judgments.

\paragraph{Gains generalise across judge families.}
As shown in Table~\ref{tab:judgesbench}, HRS consistently outperforms the strongest post-hoc calibration baselines across Qwen, Llama, and Gemma judges. It achieves lower JSD in 46 of 52 benchmark-level comparisons, including 37 statistically significant improvements. Notably, these gains hold even against an optimistic baseline that selects the best calibrator for each benchmark using test performance. This consistency across model families and sizes suggests that HRS's advantages are not specific to a particular judge architecture. Complete per-benchmark results appear in Appendix Table~\ref{tab:judgesbenchfull}, with cross-seed variability in Appendix Table~\ref{tab:seeds14}.

\subsection{Generalisation to Aggregate-Score Agreement}
\label{sec:agg}
\label{sec:transfer}

Although HRS targets distributions of individual human ratings, conventional LLM-as-a-Judge evaluation typically measures agreement with aggregated human scores. We therefore examine whether HRS generalises to this setting without additional training or an aggregate-score objective. For each ordinal benchmark, we compare the expected rating under the predicted distribution with the mean human rating using Spearman's $\rho$ and mean absolute error (MAE).

As shown in Table~\ref{tab:h2}, HRS achieves the highest mean Spearman correlation ($0.579$), substantially improving over the raw judge ($0.449$) and the strongest baseline evaluated across all criteria ($0.474$). It leads on 8 of 11 criteria, with similar gains observed for Llama (Appendix Table~\ref{tab:h2llama}), and consistently achieves lower MAE than applicable baselines (Appendix Table~\ref{tab:h2mae}). These findings demonstrate that HRS's distributional alignment also benefits conventional score-level evaluation, without explicitly optimising for aggregate-score agreement.

\begin{table}[t]
\centering\footnotesize\setlength{\tabcolsep}{5pt}
\resizebox{\columnwidth}{!}{%
\begin{tabular}{l|cc}
\hline
Method & Mean $\rho\uparrow$ & Best / total \\
\hline
Raw judge & .449 & 0/11 \\
G-Eval~\citep{geval} & .468 & 2/11 \\
LAGER~\citep{lager} & .425 & 0/11 \\
SAJA, regression~\citep{saja} & .468 & 0/11 \\
SAJA, classification & .474 & 1/11 \\
HypoEval~\citep{hypoeval} & .455 & 0/10 \\
\textbf{HRS (ours)} & \textbf{.579} & \textbf{8/11} \\
\hline
\end{tabular}}
\caption{Aggregate-score agreement among evaluated frozen-backbone methods using Qwen2.5-7B. Per-criterion correlations, MAE, and Llama-3.1-8B results appear in Appendix Tables~\ref{tab:h2full}, \ref{tab:h2mae}, and~\ref{tab:h2llama}, respectively.}
\label{tab:h2}
\end{table}

\begin{figure*}[t]
\centering
\includegraphics[width=0.8\textwidth]{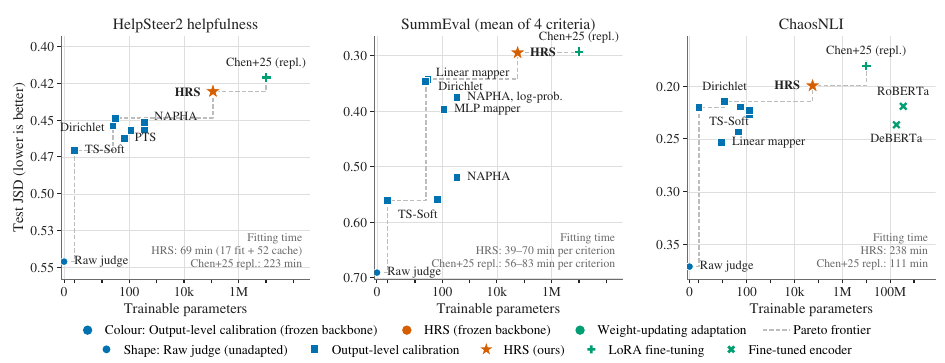}
\caption{Test JSD (lower is better) versus trainable parameters on three benchmarks (Qwen2.5-7B, seed 42). Colours and markers mark method families; dashed lines are Pareto frontiers; parameter counts exclude the judge backbone (details in Appendix \ref{app:encoder_baselines}).}
\label{fig:resource_tradeoffs}
\end{figure*}

\subsection{Ablation Study}
\label{sec:ablation}
\paragraph{Training-Loss ablation.}
We investigate the benefit of ordinal-aware supervision by replacing RPS with cross-entropy (CE), while keeping the architecture and training protocol fixed. Across four judge configurations, RPS achieves lower mean JSD in 39 of 42 ordinal benchmark comparisons, including 30 statistically significant improvements. On Qwen2.5-7B, it reduces mean JSD from $0.389$ to $0.370$, improving all eleven benchmarks. However, gains under annotation-level RPS are modest ($0.002$), suggesting that the advantage is more pronounced for distributional alignment measured by JSD. Full results are reported in Appendix Table~\ref{tab:objective}.

\paragraph{Component ablation.}
We examine the effect of oracle-based initialisation by
comparing matched rank-8 HRS variants with oracle-SVD
and random basis initialisation. On HelpSteer2 with
Qwen2.5-7B (seed 42), both achieve comparable performance
(test JSD: $0.433$ vs.\ $0.429$; annotation-level RPS:
$0.644$ vs.\ $0.645$). These results suggest that HRS
is not strongly dependent on oracle initialisation
in this setting. Rather, the oracle analysis motivates
the low-rank intervention design, while predictive
alignment is learned through input-conditioned steering
(Appendix Table~\ref{tab:components_short}).

\subsection{Trade-offs across Method Families}
\label{sec:tradeoffs}

\paragraph{Distributional alignment and parameter efficiency.}
We extend the frozen-backbone comparisons in \S\ref{sec:main} to weight-updating distributional fine-tuning, examining trade-offs in alignment quality, trainable parameters, and computational cost.
Figure~\ref{fig:resource_tradeoffs} compares output-level calibration, HRS, and weight-updating fine-tuning in alignment quality and trainable parameter requirements. While output-level calibrators use fewer parameters, they generally achieve weaker distributional alignment. Compared with LoRA-based distributional fine-tuning~\citep{chen2025distributional}, HRS achieves competitive mean JSD ($0.352$ versus $0.338$) while training substantially fewer parameters (115k versus 10.1M, approximately 88$\times$ fewer). These results position HRS as a parameter-efficient alternative to weight-updating methods, offering a favourable trade-off between distributional alignment and trainable parameter count.

\paragraph{Population-level estimation efficiency.}
We further examine whether distributional alignment improves the estimation of population-level human judgment statistics. Following \citet{fsa_asi}, we combine judge predictions with human annotations for 10\% of evaluation items and measure statistical efficiency through effective sample size (ESS). On HelpSteer2 and HelpSteer3 code, HRS improves ESS by 38-39\% relative to the classical human-only estimator, whereas the raw judge provides limited or negative gains. Empirical coverage remains close to the nominal 95\% confidence level (Appendix~\ref{app:extra}). These findings demonstrate that HRS can improve the precision of population-level estimates under limited evaluation-stage annotation budgets, beyond its benefits for instance-level distributional alignment.

\section{Conclusion}
\label{sec:conclusion}

We presented Human-Residual Steering (HRS), a representation-level approach for aligning frozen LLM judges with distributions of individual human ratings. Label-dependent oracle analyses reveal that substantial distributional discrepancies can be corrected within compact, low-dimensional hidden-state subspaces. Building on this observation, HRS learns input-conditioned interventions from individual annotations while preserving the judge's backbone parameters.

Across 14 ordinal and nominal benchmarks and multiple judge families, HRS consistently outperforms evaluated frozen-weight methods in distributional alignment and achieves the strongest aggregate-score agreement among them. Improved distributional predictions also increase the precision of downstream population-level estimates under limited evaluation-stage human annotation budgets. Although distributional fine-tuning achieves stronger alignment, HRS provides a parameter-efficient alternative with different computational and storage trade-offs. These findings establish hidden-state steering as a promising approach to human-distribution alignment, while highlighting the remaining challenges of cross-domain adaptation and preserving already-aligned predictions.

\section*{Limitations}
HRS requires access to intermediate representations
and gradients through frozen downstream layers,
limiting its applicability to API-only judges.
Its effectiveness also depends on criterion-specific
human annotations: preliminary experiments show
limited cross-benchmark transfer and mixed advantages
under small training budgets. Although HRS updates
few parameters, oracle-based initialisation and
representation caching introduce additional
computational and storage costs. Finally,
interventions can degrade some initially accurate
predictions. Improving transferability, annotation
efficiency, and prediction preservation remains
future work.

\bibliography{references}

\appendix
\section{Datasets and Evaluation Protocol}
\label{app:data_protocol}

\subsection{Benchmarks and Splits}
\label{app:benchmarks}
\begin{table*}[h]
\centering\small
\resizebox{\textwidth}{!}{%
\begin{tabular}{lllrl}
\hline
Benchmark & Label type & $K$ & Train / val.\ / test items & Raters per item \\
\hline
\multicolumn{5}{l}{\textit{14-benchmark study}} \\
SummEval, 4 dimensions \citep{summeval} & ordinal & 5 & 960 / 320 / 320 & 3 (experts) \\
Topical-Chat overall \citep{usr} & ordinal & 5 & 216 / 72 / 72 & 3 \\
Topical-Chat natural, context, engaging & ordinal & 3 & 216 / 72 / 72 & 3 \\
FLASK \citep{flask} & ordinal & 5 & 1,195 / 404 / 402 & 3 \\
HelpSteer2 helpfulness \citep{helpsteer2} & ordinal & 5 & 14,190 / 4,730 / 4,732 & 3.4 (mean) \\
HelpSteer3 code \citep{helpsteer3} & ordinal, pairwise & 6 & 5,336 / 1,730 / 1,761 & 3 \\
ChaosNLI \citep{chaosnli} & nominal & 3 & 17,986 / 6,015 / 3,113 & 5 (train, val.), 100 (test) \\
DynaSent round 2 \citep{dynasent} & nominal & 4 & 11,121 / 3,707 / 3,707 & 5 \\
Scruples Anecdotes \citep{scruples} & nominal & 5 & 8,000 / 2,437 / 2,447 & 8 (median votes) \\
\hline
\end{tabular}}
\caption{Benchmarks. All splits are by prompt, source document, conversation or instruction, so that no source appears in two splits; ChaosNLI and Scruples use fixed splits.}
\label{tab:benchmarks}
\end{table*}
\paragraph{SummEval.} 100 articles with 16 system summaries each, split by article. We use the three expert ratings only: the five crowd ratings are identical copies in 807 of 1,600 records and their means are uncorrelated with the experts' ($r=0.01$--$0.07$). Consistency and fluency are heavily skewed, with about 85\% of ratings equal to 5.
\paragraph{Topical-Chat.} We use the USR annotations \citep{usr} for 60 conversations, each with six responses and three raters, and split by conversation. We evaluate the four non-binary criteria; the two binary questions are outside the study's scope. The test split contains 72 items, resulting in wide confidence intervals.
\paragraph{FLASK.} The three human scores per row are released with Prometheus \citep{prometheus}, not with FLASK itself; they reproduce the agreement reported for FLASK (Krippendorff's $\alpha=0.48$). The judge input contains the skill-specific five-level rubric and the reference answer; splits are by instruction (169 instructions, up to three skills, four models).
\paragraph{HelpSteer2, HelpSteer3.} HelpSteer2 helpfulness has 23,652 responses to 11,825 prompts; HelpSteer3 8,827 code-preference pairs with three raters.
\paragraph{ChaosNLI.} The test split is the SNLI and MultiNLI part of ChaosNLI, 3,113 items with 100 raters each. Training and validation use the five-annotator items of the SNLI development and test sets and the MultiNLI development sets, after removing every ChaosNLI premise. No method is trained on the 100-annotator ChaosNLI test set; supervised alignment uses only the disjoint five-annotator training and validation items.
\paragraph{DynaSent.} The released development and test files of round 2 keep only sentences on which at least four of five annotators agree, which makes them unsuitable for distribution alignment. We re-split the 18,535 sentences of the training file 60/20/20.
\paragraph{Scruples.} Official splits of the Anecdotes; vote histograms with more than 100 votes are rescaled to 100 so that a few viral posts do not dominate the annotation-level scores, and training uses a random subset of 8,000 of the 27.8k training anecdotes.

\subsection{Evaluation and Statistical Protocol}
Ordinal labels have a meaningful category order; nominal labels do not. The primary metric is Jensen--Shannon distance (JSD), the square root of Jensen--Shannon divergence computed with natural logarithms. We additionally report annotation-level RPS for ordinal labels and the Brier score for nominal labels, together with aggregate-score agreement where applicable. All trainable methods use training data only; validation data or training-set cross-validation determines hyperparameters and checkpoints. For the 14-benchmark distributional and aggregate-score comparisons, uncertainty is estimated with a paired cluster bootstrap over prompts or source groups: 2,000 replicates for the three-seed Qwen2.5-7B means and 1,000 for individual judge--seed comparisons. Each three-seed replicate resamples clusters independently within each seed and then averages the seed-level scores; methods share the resampled clusters within a seed. A $^\dagger$ denotes a 95\% interval for the paired difference that excludes zero. Criteria derived from the same source items are therefore related evaluations, not independent dataset replications. In cross-judge summaries, ``best post-hoc'' is selected retrospectively for each test criterion and judge and is an optimistic comparison rather than a deployable selection rule.

\section{Baselines and Model Implementation}
\label{app:baselines_implementation}

\subsection{Baselines of the 14-Benchmark Study}
\label{app:expbaselines}
Baseline comparisons use the same base judge, data splits, and bootstrap procedure as \method. Below, we describe each implementation, including method-specific prompts and departures from the original procedure.
\paragraph{TS-Soft \citep{dogah2026}.} Our implementation fits one temperature by L-BFGS to the Brier score against the human histograms. The paper fits it on half of the soft-label test set as an oracle; we fit it on the training split. The paper's public record links code and per-seed results.
\paragraph{PTS \citep{pts}.} The temperature is the absolute value of an MLP (two hidden layers of five units) on the sorted logits, trained as in the paper (squared error, Adam at $5\times10^{-5}$, batch 1,000, 100,000 steps, no early stopping), with the annotator histogram as the target instead of a one-hot label.
\paragraph{Dirichlet calibration \citep{kull2019}.} $\mathrm{softmax}(W\ln q+b)$ fitted by the log-loss over individual annotations with the ODIR regulariser; its two weights are chosen by five-fold cross-validated log-loss on the training split and the five fold models are averaged, as in the paper.
\paragraph{Linear mapper \citep{vandenburg2025}.} Ridge least squares ($\lambda=10^{-6}$) from the one-hot label of the judge to the human labels; a test item receives the row of the fitted matrix of the judge's argmax label. The paper fits it on 100 items; the main table uses the full training split, and the full tables add the 100-item fit and an extension on the judge's probabilities that the authors mention but do not test.
\paragraph{MLP mapper and NAPHA \citep{napha}.} The single alignment model is one $K\to2K\to K$ MLP trained with KL to the histogram; NAPHA routes items to three such MLPs by the entropy tercile of the judge's distribution. Optimiser settings are not given in the paper; we use Adam (lr $10^{-3}$, weight decay $10^{-2}$) with early stopping on validation, and choose between raw and smoothed targets on validation, both in the baseline's favour. The paper feeds verbalized soft labels of a black-box LLM and names logits as future work; our input is the judge's first-token distribution, as probabilities or as log-probabilities. No code is available.
\paragraph{Replication of \citet{chen2025distributional} (distributional LoRA fine-tuning).} The loss is $0.8\,\mathrm{KL}(p'\|q)+0.2\,\mathrm{CE}(q,\text{majority})$, with $p'$ the worst case in an $\ell_2$ ball of radius $0.25$ around the histogram (five projected-gradient steps, step size $0.05$), optimised with AdamW at $5\times10^{-5}$. The LoRA shape is not given in the paper; we use rank 16 on the query, key, value and output projections of every block. We keep the validation-best checkpoint instead of training for two epochs without validation, and raise the number of epochs on the smallest training sets (at least 600 steps, at most 20 epochs). No code is available, so this is a re-implementation from the paper's description. It is the row, column, or marker labelled \emph{\citet{chen2025distributional}} in every table of the 14-benchmark study, in Figure~\ref{fig:resource_tradeoffs}, and in the transfer and label-budget results (Table~\ref{tab:generalisation}).
\paragraph{G-Eval \citep{geval}.} Token-probability mode: the form line starts the reply and the next-token distribution over the score digits is read. We use the released prompts for SummEval coherence, consistency and relevance, and those of \citet{chiang2023closer} for Topical-Chat. SummEval fluency is scored on the benchmark's 1--5 scale (the released prompt uses 1--3); for the remaining criteria the evaluation steps are written by the judge in the G-Eval layout.
\paragraph{LAGER \citep{lager}.} As in the released code: the unembedding is applied to the hidden state of every layer, and the layer weights are uniform or tuned with cross-entropy plus MSE (Adam, lr $0.01$, batch 4, two epochs) on 1,000 labels. The 1,000 tuning items are in-domain training items; the paper tunes once on HelpSteer. The full tables also list an in-domain RPS-tuned five-layer variant.
\paragraph{SAJA \citep{saja}.} The released broad rubric (seven dimensions and an overall score, one JSON call per item); the head family and its hyper-parameters are chosen by five-fold grouped cross-validation, for regression on the mean rating or classification of the aggregated label, on at most 1,500 labelled items. Missing rubric scores are imputed with the corresponding training-set median (3 if the dimension has no parsed training scores), including items with no parseable scores. The fine-grained 28-score rubric is not run.
\paragraph{HypoEval \citep{hypoeval}.} The scoring prompt with five hypotheses and the mean of five chain-of-thought scores. On SummEval we use the authors' released hypothesis sets; elsewhere the bank is generated by the judge from 30 labelled training items with the released generation prompt, and the five hypotheses with the highest Pearson correlation are kept. The literature half of the bank and the refinement loop are not reproduced. Unparseable replies are dropped from the mean (at most 2 of 1,600 on SummEval, 2.5\% on HelpSteer2); if no hypothesis yields a score, the implementation falls back to the raw judge's expected score. The method does not apply to the pairwise HelpSteer3 task.
\paragraph{Scalar-only rows.} The SAJA regression head and HypoEval output a scalar; their probability-based columns are left blank. Correlations are computed at the dataset level for every row, whereas G-Eval and HypoEval report summary-level correlations on SummEval, so our numbers are not comparable with the published ones.
\paragraph{Annotation budgets in the aggregate study.} The aggregate-score comparison is not annotation-budget matched. LAGER uses 1,000 labelled training items, SAJA at most 1,500, and HypoEval 30 outside SummEval (where released hypotheses are used), whereas \method uses the full benchmark training split. We therefore treat Tables~\ref{tab:h2full} and~\ref{tab:h2llama} as a secondary generalisation study, not evidence of superiority at equal annotation cost.
\paragraph{\method on the 14 benchmarks.} Rank 8 (16 on HelpSteer2), linear router with random initialisation, trainable $V$ initialised with the SVD of training-item oracle corrections, AdamW at $5\times10^{-4}$, $\lambda_a=10^{-4}$. Oracle construction and SVD initialisation are part of every reported primary run; they are not optional for reproducing those rows. A targeted random-initialisation control is reported separately in Table~\ref{tab:components_short}. Short inputs are trained in exact mode with the batch size of the judge (8; 4 for Gemma-2-9B and Qwen2.5-14B), HelpSteer2, HelpSteer3 and Scruples through a partial key--value store with batches of 32. The cap on the number of epochs and the patience are 20 / 6 on ChaosNLI and DynaSent, 40 / 8 on SummEval and FLASK, 60 / 10 on Topical-Chat, and 30 / 5 (HelpSteer2, HelpSteer3) or 30 / 10 (Scruples) with a store. Every run selects the epoch on up to 4,000 validation items and the scale $\alpha\in\{0.5,0.75,1,1.25,1.5\}$ on the full validation split. The steering block is the argmin of the validation oracle score for Qwen2.5-7B (a tolerance of $0.005$ with the deepest candidate on Scruples) and, for the other judges, the deepest of the five candidate depths within $0.005$ of the best; the Gemma-2 grid adds the depths $0.9L$ and $0.96L$ (blocks 37 and 40) because its cached suffix is four times larger per block than that of Qwen2.5-7B. Table~\ref{tab:hrsettings} lists the resulting settings.

\subsection{ChaosNLI Encoder Comparisons}
\label{app:encoder_baselines}
The encoder points in Figure~\ref{fig:resource_tradeoffs} are our
fine-tuning runs, starting from
\texttt{FacebookAI/roberta-large-mnli} \citep{roberta} and
\texttt{microsoft/deberta-v3-base} \citep{debertav3}.
Both use the same 17,986 five-annotator training items as the other
ChaosNLI comparisons, with ChaosNLI premises removed. We update all
encoder weights using annotation-level multiclass Brier loss for four
epochs with AdamW, at learning rates $10^{-5}$ and $2\times10^{-5}$,
respectively, and select the checkpoint by validation Brier score.
The resulting models are evaluated on the disjoint 100-annotator
ChaosNLI test set. These are local comparisons under our data and
training protocol, rather than results reported by \citet{chaosnli}.

\section{Complete Distributional Evaluation Results}
\label{app:distributional_results}

\subsection{Detailed Cross-Judge Results}
\label{app:crossjudge}
Table~\ref{tab:judgesbenchfull} reports every per-benchmark value behind the
cross-judge summary in \S\ref{sec:main}. All cells within the specified run
coverage are complete; the Qwen2.5-14B exclusions are explicit in the caption.

\begin{table*}[t]
\centering\scriptsize\setlength{\tabcolsep}{3pt}
\begin{tabular}{@{}l*{14}{c}c@{}}
\toprule
 & \multicolumn{4}{c}{\fontsize{5}{6}\selectfont SummEval} & \multicolumn{4}{c}{\fontsize{5}{6}\selectfont Topical-Chat} & & & & & & & \\
\cmidrule(lr){2-5}\cmidrule(lr){6-9}
Method & Coh & Con & Flu & Rel & Ovr & Nat & Ctx & Eng & FLASK & HS2 & HS3 & CNLI & DynaS & Scrup & Mean \\
\midrule
\multicolumn{16}{l}{\textit{Qwen2.5-7B, mean of seeds}} \\
Raw judge & .609$^{\dagger}$ & .772$^{\dagger}$ & .786$^{\dagger}$ & .590$^{\dagger}$ & .594$^{\dagger}$ & .492$^{\dagger}$ & .500$^{\dagger}$ & .411$^{\dagger}$ & .572$^{\dagger}$ & .542$^{\dagger}$ & .641$^{\dagger}$ & \multicolumn{1}{|c}{.371$^{\dagger}$} & .342$^{\dagger}$ & .683$^{\dagger}$ & \multicolumn{1}{|c}{.565} \\
Best post-hoc & .473$^{\dagger}$ & .209$^{\dagger}$ & .288$^{\dagger}$ & .401$^{\dagger}$ & .473$^{\dagger}$ & \textbf{.376} & .352 & .345$^{\dagger}$ & .425$^{\dagger}$ & .448$^{\dagger}$ & .576$^{\dagger}$ & \multicolumn{1}{|c}{.214$^{\dagger}$} & .324$^{\dagger}$ & .375$^{\dagger}$ & \multicolumn{1}{|c}{.377} \\
\textbf{HRS (ours)} & \textbf{.414} & \textbf{.159} & \textbf{.241} & \textbf{.383} & \textbf{.440} & .379 & \textbf{.349} & \textbf{.321} & \textbf{.410} & \textbf{.432} & \textbf{.546} & \multicolumn{1}{|c}{\textbf{.198}} & \textbf{.302} & \textbf{.348} & \multicolumn{1}{|c}{\textbf{.352}} \\
\midrule
\multicolumn{16}{l}{\textit{Llama-3.1-8B}} \\
Raw judge & .494$^{\dagger}$ & .684$^{\dagger}$ & .707$^{\dagger}$ & .448$^{\dagger}$ & .562$^{\dagger}$ & .396 & .387 & .384 & .523$^{\dagger}$ & .471$^{\dagger}$ & .622$^{\dagger}$ & \multicolumn{1}{|c}{.349$^{\dagger}$} & .353$^{\dagger}$ & .478$^{\dagger}$ & \multicolumn{1}{|c}{.490} \\
Best post-hoc & .456$^{\dagger}$ & .233$^{\dagger}$ & .302$^{\dagger}$ & .395 & .473 & \textbf{.366} & .379 & \textbf{.374} & .446$^{\dagger}$ & .449$^{\dagger}$ & .601$^{\dagger}$ & \multicolumn{1}{|c}{.250$^{\dagger}$} & .325 & .368$^{\dagger}$ & \multicolumn{1}{|c}{.387} \\
\textbf{HRS (ours)} & \textbf{.435} & \textbf{.136} & \textbf{.264} & \textbf{.392} & \textbf{.464} & .387 & \textbf{.373} & .376 & \textbf{.392} & \textbf{.435} & \textbf{.580} & \multicolumn{1}{|c}{\textbf{.227}} & \textbf{.324} & \textbf{.347} & \multicolumn{1}{|c}{\textbf{.367}} \\
\midrule
\multicolumn{16}{l}{\textit{Gemma-2-9B}} \\
Raw judge & .615$^{\dagger}$ & .792$^{\dagger}$ & .803$^{\dagger}$ & .665$^{\dagger}$ & .579$^{\dagger}$ & .428$^{\dagger}$ & .447$^{\dagger}$ & .367 & .575$^{\dagger}$ & .614$^{\dagger}$ & .622$^{\dagger}$ & \multicolumn{1}{|c}{.368$^{\dagger}$} & .325 & .574$^{\dagger}$ & \multicolumn{1}{|c}{.555} \\
Best post-hoc & .421$^{\dagger}$ & .216$^{\dagger}$ & .262$^{\dagger}$ & .387$^{\dagger}$ & .449 & .353$^{\dagger}$ & .362$^{\dagger}$ & .339 & .429$^{\dagger}$ & .438$^{\dagger}$ & .588$^{\dagger}$ & \multicolumn{1}{|c}{.220$^{\dagger}$} & \textbf{.316} & .355$^{\dagger}$ & \multicolumn{1}{|c}{.367} \\
\textbf{HRS (ours)} & \textbf{.406} & .237 & \textbf{.218} & \textbf{.377} & \textbf{.433} & \textbf{.325} & \textbf{.339} & \textbf{.320} & \textbf{.404} & \textbf{.426} & \textbf{.552} & \multicolumn{1}{|c}{\textbf{.206}} & .318 & \textbf{.338} & \multicolumn{1}{|c}{\textbf{.350}} \\
\midrule
\multicolumn{16}{l}{\textit{Qwen2.5-14B}} \\
Raw judge & .652$^{\dagger}$ & .737$^{\dagger}$ & .810$^{\dagger}$ & .626$^{\dagger}$ & .557$^{\dagger}$ & .396 & .482$^{\dagger}$ & .394$^{\dagger}$ & .537$^{\dagger}$ & -- & -- & \multicolumn{1}{|c}{--} & .320$^{\dagger}$ & -- & \multicolumn{1}{|c}{.551} \\
Best post-hoc & .444$^{\dagger}$ & .189$^{\dagger}$ & .275$^{\dagger}$ & .388$^{\dagger}$ & .459 & .352 & .367 & .350$^{\dagger}$ & .423$^{\dagger}$ & -- & -- & \multicolumn{1}{|c}{--} & .317$^{\dagger}$ & -- & \multicolumn{1}{|c}{.356} \\
\textbf{HRS (ours)} & \textbf{.401} & \textbf{.171} & \textbf{.238} & \textbf{.378} & .465 & \textbf{.340} & \textbf{.359} & \textbf{.312} & \textbf{.401} & -- & -- & \multicolumn{1}{|c}{--} & \textbf{.295} & -- & \multicolumn{1}{|c}{\textbf{.336}} \\
\bottomrule
\end{tabular}
\caption{Complete cross-judge distributional comparison underlying Table~\ref{tab:judgesbench}: test JSD by benchmark. Qwen2.5-7B values are the mean of seeds 42, 13, and 7; other judges use seed 42. Best post-hoc is selected separately for each criterion and judge by test JSD, giving an optimistic retrospective envelope. \textbf{Bold}: lower of the two trained rows; $^\dagger$: significant paired difference from HRS in either direction. --: HelpSteer2, HelpSteer3, ChaosNLI, and Scruples were not evaluated in the Qwen2.5-14B configuration; they are outside the run design, not pending results. Its mean covers the ten evaluated benchmarks. Objective and proper-score comparisons appear in Tables~\ref{tab:objective} and~\ref{tab:h1rps}.}
\label{tab:judgesbenchfull}
\end{table*}

\subsection{Annotation-Level Proper Scores}
\label{app:scores}
This section supplements the main JSD results with proper scores evaluated on individual annotations: RPS for ordinal ratings and the Brier score for nominal labels. These scores assess predictions against observed ratings, whereas JSD compares them with the empirical histogram. Both estimates remain sensitive to the available annotations; scoring individual ratings does not remove sampling uncertainty. Model-selection procedures are specified in Appendix~\ref{app:expbaselines}.

Table~\ref{tab:h1rps} reports these complementary scores and the training-objective variant. Across three seeds, \method achieves a lower proper score than every post-hoc baseline on 12 of 14 benchmarks, with 8 significant improvements and no significant degradation. The cross-benchmark means are $0.466$ for \method, $0.495$ for Dirichlet calibration, and $0.989$ for the raw judge. These summaries combine ordinal RPS and nominal Brier scores; the main paper uses JSD for a common cross-benchmark scale. The exceptions are Topical-Chat naturalness and context, where \method is worse by $0.004$ and $0.012$, respectively.

The relative performance of training objectives depends on the evaluation metric. On ordinal benchmarks, RPS training wins 8 comparisons to CE's 3 under annotation-level RPS, with a mean advantage of $0.002$. Under JSD, it wins all 11 comparisons, including 10 significant improvements (Table~\ref{tab:objective}).

\paragraph{RPS versus CE across judges.}
Table~\ref{tab:objective} covers all eleven ordinal benchmarks for Qwen,
Llama, and Gemma, plus the nine included in the Qwen2.5-14B configuration.
On the primary judge, RPS lowers mean JSD from $0.389$ to $0.370$,
improving all eleven benchmarks, ten significantly. It also lowers the
ordinal mean for every other judge: $0.403\to0.385$ for Llama,
$0.387\to0.367$ for Gemma, and $0.357\to0.341$ for Qwen2.5-14B.
Overall, RPS wins 39 of 42 ordinal comparisons, with 30 significant
improvements. These are descriptive comparisons with shared datasets,
not independent trials.

The consistent JSD advantage supports accounting for ordinal structure
when learning human rating distributions. It is not a universal advantage
on every metric: the primary judge's mean annotation-level RPS improves by
only $0.002$. We therefore conclude
that RPS is the stronger default for ordinal \emph{distributional alignment},
rather than claiming superiority on all evaluation targets. For the three
nominal benchmarks, label order is undefined and primary HRS uses CE;
their complete results are included in Table~\ref{tab:objective}, but they
are not counted as RPS--CE ablations.

The nominal objective also involves a metric-dependent choice. In the
seed-42 controls, replacing CE with annotation-level Brier loss lowers JSD
on ChaosNLI ($.200\to.196$), DynaSent ($.292\to.267$), and Scruples
($.345\to.341$). It improves annotation-level Brier only on DynaSent
($.455\to.446$), however, and is neutral or worse on the other two tasks.
Thus CE is the primary nominal recipe, not a claim that it dominates every
task-appropriate proper loss.

\begin{table*}[t]
\centering\scriptsize\setlength{\tabcolsep}{5pt}
\begin{tabular}{l|cc|cc|cc|cc}
\hline
 & \multicolumn{2}{c|}{Qwen2.5-7B} & \multicolumn{2}{c|}{Llama-3.1-8B} & \multicolumn{2}{c|}{Gemma-2-9B} & \multicolumn{2}{c}{Qwen2.5-14B} \\
Benchmark & RPS & CE & RPS & CE & RPS & CE & RPS & CE \\
\hline
\multicolumn{9}{l}{\textit{Ordinal labels: only the training objective changes}} \\
SummEval coherence & \textbf{.414} & .442$^{\dagger}$ & \textbf{.435} & .468$^{\dagger}$ & \textbf{.406} & .432$^{\dagger}$ & \textbf{.401} & .428$^{\dagger}$ \\
SummEval consistency & \textbf{.159} & .180$^{\dagger}$ & \textbf{.136} & .195$^{\dagger}$ & .237 & \textbf{.178}$^{\dagger}$ & \textbf{.171} & .172 \\
SummEval fluency & \textbf{.241} & .271$^{\dagger}$ & \textbf{.264} & .282$^{\dagger}$ & \textbf{.218} & .302$^{\dagger}$ & \textbf{.238} & .239 \\
SummEval relevance & \textbf{.383} & .398$^{\dagger}$ & \textbf{.392} & .400$^{\dagger}$ & \textbf{.377} & .391$^{\dagger}$ & \textbf{.378} & .395$^{\dagger}$ \\
Topical-Chat overall & \textbf{.440} & .457$^{\dagger}$ & \textbf{.464} & .476 & \textbf{.433} & .459$^{\dagger}$ & .465 & \textbf{.459} \\
Topical-Chat naturalness & \textbf{.379} & .381 & \textbf{.387} & .399 & \textbf{.325} & .359$^{\dagger}$ & \textbf{.340} & .340 \\
Topical-Chat context & \textbf{.349} & .374$^{\dagger}$ & \textbf{.373} & .383 & \textbf{.339} & .385$^{\dagger}$ & \textbf{.359} & .383$^{\dagger}$ \\
Topical-Chat engaging & \textbf{.321} & .359$^{\dagger}$ & \textbf{.376} & .389 & \textbf{.320} & .353$^{\dagger}$ & \textbf{.312} & .369$^{\dagger}$ \\
FLASK & \textbf{.410} & .426$^{\dagger}$ & \textbf{.392} & .427$^{\dagger}$ & \textbf{.404} & .405 & \textbf{.401} & .427$^{\dagger}$ \\
HelpSteer2 & \textbf{.432} & .440$^{\dagger}$ & \textbf{.435} & .449$^{\dagger}$ & \textbf{.426} & .441$^{\dagger}$ & -- & -- \\
HelpSteer3 code & \textbf{.546} & .548$^{\dagger}$ & .580 & \textbf{.571}$^{\dagger}$ & \textbf{.552} & .556$^{\dagger}$ & -- & -- \\
\hline
Ordinal mean & \textbf{.370} & .389 & \textbf{.385} & .403 & \textbf{.367} & .387 & \textbf{.341} & .357 \\
Ordinal JSD wins & 11/11 & 0/11 & 10/11 & 1/11 & 10/11 & 1/11 & 8/9 & 1/9 \\
\hline
\multicolumn{9}{l}{\textit{Nominal labels: CE is the primary objective; no RPS comparison}} \\
ChaosNLI & \multicolumn{2}{c|}{.198} & \multicolumn{2}{c|}{.227} & \multicolumn{2}{c|}{.206} & \multicolumn{2}{c}{--} \\
DynaSent R2 & \multicolumn{2}{c|}{.302} & \multicolumn{2}{c|}{.324} & \multicolumn{2}{c|}{.318} & \multicolumn{2}{c}{.295} \\
Scruples Anecdotes & \multicolumn{2}{c|}{.348} & \multicolumn{2}{c|}{.347} & \multicolumn{2}{c|}{.338} & \multicolumn{2}{c}{--} \\
\hline
\end{tabular}
\caption{Training-objective ablation across the benchmark suite: test JSD (lower is better). The RPS and CE columns denote training losses, not evaluation metrics. Architecture, data, and validation protocol are held fixed within each pair. Qwen2.5-7B values are the mean of seeds; other judges use seed 42. \textbf{Bold}: lower unrounded JSD within an ordinal pair. $^\dagger$: significant paired difference from RPS in either direction. Nominal rows report primary CE results only because RPS requires ordered labels. --: not evaluated in the Qwen2.5-14B configuration. The primary-judge proper-score counterpart is Table~\ref{tab:h1rps}.}
\label{tab:objective}
\end{table*}

\begin{table*}[h]
\centering\scriptsize\setlength{\tabcolsep}{3.5pt}
\begin{tabular}{@{}l*{14}{c}c@{}}
\toprule
 & \multicolumn{11}{c}{\fontsize{5}{6}\selectfont Ordinal benchmarks: annotation RPS $\downarrow$} & \multicolumn{3}{c}{\fontsize{5}{6}\selectfont Nominal: Brier $\downarrow$} & \\
\cmidrule(lr){2-12}\cmidrule(lr){13-15}
 & \multicolumn{4}{c}{\fontsize{5}{6}\selectfont SummEval} & \multicolumn{4}{c}{\fontsize{5}{6}\selectfont Topical-Chat} &  &  &  &  &  &  & \\
\cmidrule(lr){2-5}\cmidrule(lr){6-9}
Method & Coh & Con & Flu & Rel & Ovr & Nat & Ctx & Eng & FLASK & HS2 & HS3 & CNLI & DynaS & Scrup & Avg. \\
\midrule
Raw judge & 1.124$^{\dagger}$ & 1.093$^{\dagger}$ & 1.618$^{\dagger}$ & .939$^{\dagger}$ & .984$^{\dagger}$ & .688$^{\dagger}$ & .644$^{\dagger}$ & .505$^{\dagger}$ & .931$^{\dagger}$ & .987$^{\dagger}$ & 1.428$^{\dagger}$ & \multicolumn{1}{|c}{.769$^{\dagger}$} & .721$^{\dagger}$ & 1.415$^{\dagger}$ & \multicolumn{1}{|c}{.989} \\
TS-Soft & .745$^{\dagger}$ & .974$^{\dagger}$ & 1.086$^{\dagger}$ & .681$^{\dagger}$ & .721$^{\dagger}$ & .459$^{\dagger}$ & .446$^{\dagger}$ & .373$^{\dagger}$ & .761$^{\dagger}$ & .708$^{\dagger}$ & .839$^{\dagger}$ & \multicolumn{1}{|c}{.570$^{\dagger}$} & .525$^{\dagger}$ & .794$^{\dagger}$ & \multicolumn{1}{|c}{.692} \\
PTS & .745$^{\dagger}$ & .931$^{\dagger}$ & 1.086$^{\dagger}$ & .672$^{\dagger}$ & .718$^{\dagger}$ & .445$^{\dagger}$ & .430$^{\dagger}$ & .372$^{\dagger}$ & .753$^{\dagger}$ & .700$^{\dagger}$ & .838$^{\dagger}$ & \multicolumn{1}{|c}{.564$^{\dagger}$} & .519$^{\dagger}$ & .788$^{\dagger}$ & \multicolumn{1}{|c}{.683} \\
Dirichlet (ODIR) & .655$^{\dagger}$ & .182$^{\dagger}$ & .247$^{\dagger}$ & .479 & .666 & \textbf{.384} & \textbf{.340} & .324 & .552 & .685$^{\dagger}$ & .763$^{\dagger}$ & \multicolumn{1}{|c}{.566$^{\dagger}$} & .488$^{\dagger}$ & .593$^{\dagger}$ & \multicolumn{1}{|c}{.495} \\
Linear mapper & .664$^{\dagger}$ & .187$^{\dagger}$ & .264$^{\dagger}$ & .488$^{\dagger}$ & .684 & .408 & .355 & .356$^{\dagger}$ & .568 & .696$^{\dagger}$ & .851$^{\dagger}$ & \multicolumn{1}{|c}{.627$^{\dagger}$} & .548$^{\dagger}$ & .597$^{\dagger}$ & \multicolumn{1}{|c}{.521} \\
MLP mapper & .669$^{\dagger}$ & .252$^{\dagger}$ & .310$^{\dagger}$ & .496$^{\dagger}$ & .753$^{\dagger}$ & .425$^{\dagger}$ & .385$^{\dagger}$ & .398$^{\dagger}$ & .574$^{\dagger}$ & .699$^{\dagger}$ & .845$^{\dagger}$ & \multicolumn{1}{|c}{.602$^{\dagger}$} & .535$^{\dagger}$ & .598$^{\dagger}$ & \multicolumn{1}{|c}{.539} \\
NAPHA & .695$^{\dagger}$ & .761$^{\dagger}$ & .739$^{\dagger}$ & .569$^{\dagger}$ & .757$^{\dagger}$ & .429$^{\dagger}$ & .390$^{\dagger}$ & .406$^{\dagger}$ & .703$^{\dagger}$ & .698$^{\dagger}$ & .842$^{\dagger}$ & \multicolumn{1}{|c}{.562$^{\dagger}$} & .515$^{\dagger}$ & .598$^{\dagger}$ & \multicolumn{1}{|c}{.619} \\
NAPHA (log-prob.) & .673$^{\dagger}$ & .197$^{\dagger}$ & .261$^{\dagger}$ & .493$^{\dagger}$ & .701$^{\dagger}$ & .425$^{\dagger}$ & .369 & .363$^{\dagger}$ & .587$^{\dagger}$ & .690$^{\dagger}$ & .779$^{\dagger}$ & \multicolumn{1}{|c}{.564$^{\dagger}$} & .493$^{\dagger}$ & .598$^{\dagger}$ & \multicolumn{1}{|c}{.514} \\
\midrule
\textbf{\method, CDF loss} & \textbf{.554} & \textbf{.162} & \textbf{.220} & .467 & .645 & .389 & .352 & \textbf{.318} & \textbf{.551} & \textbf{.638} & .694 & \multicolumn{1}{|c}{--} & -- & -- & \multicolumn{1}{|c}{\textbf{.466}} \\
\textbf{\method, CE loss} & .580$^{\dagger}$ & .170 & .221 & \textbf{.467} & \textbf{.610}$^{\dagger}$ & .395 & .356 & .329 & .554 & .649$^{\dagger}$ & \textbf{.685}$^{\dagger}$ & \multicolumn{1}{|c}{\textbf{.535}} & \textbf{.450} & \textbf{.553} & \multicolumn{1}{|c}{.468} \\
\bottomrule
\end{tabular}
\caption{Extended counterpart to Table~\ref{tab:h1} under the annotation-level proper score: RPS on the ordinal and Brier score on the nominal benchmarks, mean over the three seeds of the Qwen2.5-7B study (lower is better). The two HRS objective rows are retained here as an ablation. Avg.: unweighted mean over the 14 benchmarks; the CDF row uses the CE values on the three nominal benchmarks, where the CDF loss is undefined. \textbf{Bold}: best row; $^\dagger$: the paired difference of the three-seed mean to the primary \method row is significant in either direction under a prompt-level bootstrap drawn within each seed.}
\label{tab:h1rps}
\end{table*}

\subsection{Complete Aggregate-Score Results}
\label{app:aggregate_details}

\begin{table*}[h]
\centering\scriptsize\setlength{\tabcolsep}{3.2pt}
\begin{tabular}{@{}l*{11}{c}c@{}}
\toprule
 & \multicolumn{4}{c}{\fontsize{5}{6}\selectfont SummEval} & \multicolumn{4}{c}{\fontsize{5}{6}\selectfont Topical-Chat} &  &  &  & \\
\cmidrule(lr){2-5}\cmidrule(lr){6-9}
Method & Coh & Con & Flu & Rel & Ovr & Nat & Ctx & Eng & FLASK & HS2 & HS3 & Avg. \\
\midrule
Raw judge & .407$^{\dagger}$ & .411$^{\dagger}$ & .350$^{\dagger}$ & .518 & .548$^{\dagger}$ & .485 & .614 & .685$^{\dagger}$ & .452$^{\dagger}$ & .298$^{\dagger}$ & .168$^{\dagger}$ & \multicolumn{1}{|c}{.449} \\
G-Eval~\citep{geval} & .436$^{\dagger}$ & \underline{.474} & .385$^{\dagger}$ & .500 & .488 & .448 & .517 & .730 & \underline{.535} & .338$^{\dagger}$ & .301$^{\dagger}$ & \multicolumn{1}{|c}{.468} \\
LAGER~\citep{lager} & .379$^{\dagger}$ & .423$^{\dagger}$ & -.115$^{\dagger}$ & .513 & .638 & .546 & .590 & .733 & .450$^{\dagger}$ & .329$^{\dagger}$ & .192$^{\dagger}$ & \multicolumn{1}{|c}{.425} \\
SAJA, regression~\citep{saja} & .516$^{\dagger}$ & .459 & .368 & .453$^{\dagger}$ & .638 & .566 & .433 & .632 & .420 & .356$^{\dagger}$ & .303$^{\dagger}$ & \multicolumn{1}{|c}{.468} \\
SAJA, classification & .486$^{\dagger}$ & .469 & .384 & .466 & .614 & \underline{.614} & .455 & .664 & .404$^{\dagger}$ & .355$^{\dagger}$ & .301$^{\dagger}$ & \multicolumn{1}{|c}{.474} \\
HypoEval~\citep{hypoeval} & .437$^{\dagger}$ & .390$^{\dagger}$ & .347$^{\dagger}$ & .503 & .621 & .430 & .465 & .668$^{\dagger}$ & .410$^{\dagger}$ & .279$^{\dagger}$ & -- & \multicolumn{1}{|c}{.455} \\
\midrule
\textbf{\method (ours)} & \underline{.666} & .473 & \underline{.470} & \underline{.562} & \underline{.697} & .584 & \underline{.618} & \textbf{\underline{.794}} & .498 & \underline{.498} & \underline{.505} & \multicolumn{1}{|c}{\underline{.579}} \\
\midrule
\citet{chen2025distributional}, replication & \textbf{.668} & \textbf{.517}$^{\dagger}$ & \textbf{.531} & \textbf{.608}$^{\dagger}$ & \textbf{.775}$^{\dagger}$ & \textbf{.645} & \textbf{.659} & .774 & \textbf{.565}$^{\dagger}$ & \textbf{.557}$^{\dagger}$ & \textbf{.632}$^{\dagger}$ & \multicolumn{1}{|c}{\textbf{.630}} \\
\bottomrule
\end{tabular}
\caption{Per-criterion aggregate-score agreement: Spearman $\rho$ between each expected score and the mean human rating on the 11 ordinal benchmarks (Qwen2.5-7B, seed 42). Avg.: mean over the available criteria (HypoEval: ten). The \citet{chen2025distributional} row is our replication of their distributional LoRA fine-tuning. \textbf{Bold}: best row; \underline{underlined}: best frozen-backbone row; $^\dagger$: significant paired difference to HRS.}
\label{tab:h2full}
\label{tab:h2ft}
\end{table*}

\begin{table*}[h]
\centering\scriptsize\setlength{\tabcolsep}{3.2pt}
\begin{tabular}{@{}l*{11}{c}c@{}}
\toprule
 & \multicolumn{4}{c}{\fontsize{5}{6}\selectfont SummEval} & \multicolumn{4}{c}{\fontsize{5}{6}\selectfont Topical-Chat} &  &  &  & \\
\cmidrule(lr){2-5}\cmidrule(lr){6-9}
Method & Coh & Con & Flu & Rel & Ovr & Nat & Ctx & Eng & FLASK & HS2 & HS3 & Avg. \\
\midrule
Raw judge & .277$^{\dagger}$ & .281$^{\dagger}$ & .226$^{\dagger}$ & .329 & .493$^{\dagger}$ & .439 & .480 & .570 & .392$^{\dagger}$ & .273$^{\dagger}$ & .002$^{\dagger}$ & \multicolumn{1}{|c}{.342} \\
G-Eval~\citep{geval} & .430$^{\dagger}$ & .393 & .184$^{\dagger}$ & .487 & \underline{.629} & .552 & .535 & .502 & .557 & .310$^{\dagger}$ & .350 & \multicolumn{1}{|c}{.448} \\
LAGER~\citep{lager} & .322$^{\dagger}$ & .325$^{\dagger}$ & .215$^{\dagger}$ & .411 & .510 & .394 & .462 & .578$^{\dagger}$ & .424$^{\dagger}$ & .284$^{\dagger}$ & .018$^{\dagger}$ & \multicolumn{1}{|c}{.358} \\
SAJA, regression~\citep{saja} & .363$^{\dagger}$ & .371$^{\dagger}$ & .316$^{\dagger}$ & .364$^{\dagger}$ & .593 & .517 & \underline{.577} & .578 & .338$^{\dagger}$ & .254$^{\dagger}$ & .271$^{\dagger}$ & \multicolumn{1}{|c}{.413} \\
SAJA, classification & .348$^{\dagger}$ & .337$^{\dagger}$ & .158$^{\dagger}$ & .346$^{\dagger}$ & .604 & \underline{.576}$^{\dagger}$ & .530 & .579 & .406$^{\dagger}$ & .255$^{\dagger}$ & .228$^{\dagger}$ & \multicolumn{1}{|c}{.397} \\
HypoEval~\citep{hypoeval} & .361$^{\dagger}$ & .414 & .267$^{\dagger}$ & .413 & .504 & .406 & .453 & .579 & .531$^{\dagger}$ & .275$^{\dagger}$ & -- & \multicolumn{1}{|c}{.420} \\
\midrule
\textbf{\method (ours)} & \underline{.575} & \underline{.451} & \underline{.414} & \underline{.518} & .620 & .369 & .521 & \underline{.615} & \textbf{\underline{.618}} & \underline{.458} & \underline{.362} & \multicolumn{1}{|c}{\underline{.502}} \\
\midrule
\citet{chen2025distributional}, replication & \textbf{.695}$^{\dagger}$ & \textbf{.516}$^{\dagger}$ & \textbf{.606}$^{\dagger}$ & \textbf{.583} & \textbf{.763}$^{\dagger}$ & \textbf{.622}$^{\dagger}$ & \textbf{.662}$^{\dagger}$ & \textbf{.739}$^{\dagger}$ & .594 & \textbf{.571}$^{\dagger}$ & \textbf{.628}$^{\dagger}$ & \multicolumn{1}{|c}{\textbf{.634}} \\
\bottomrule
\end{tabular}
\caption{Per-criterion aggregate-score agreement with the Llama-3.1-8B judge: Spearman $\rho$ between each expected score and the mean human rating. Avg.: mean over the available criteria (HypoEval: ten). The \citet{chen2025distributional} row is our replication of their distributional LoRA fine-tuning. \textbf{Bold}: best row; \underline{underlined}: best frozen-backbone row; $^\dagger$: significant paired difference to HRS.}
\label{tab:h2llama}
\end{table*}

\subsection{Additional Aggregate Results}
\label{app:expfull}
Table~\ref{tab:wl} summarises benchmark-level wins and losses across evaluation metrics, while Table~\ref{tab:h2mae} reports the complementary aggregate-score MAE results.
\begin{table*}[h]
\centering\scriptsize\setlength{\tabcolsep}{3.5pt}
\begin{tabular}{lcccc}
\hline
\multicolumn{5}{l}{\textit{(a) Distribution alignment, 14 benchmarks}} \\
HRS vs. & $\rho$ & KL & RPS/Brier & JSD \\
\hline
Raw judge & 14--0 (11/0) & 14--0 (14/0) & 14--0 (14/0) & 14--0 (13/0) \\
TS-Soft & 13--1 (8/0) & 13--1 (10/0) & 14--0 (13/0) & 14--0 (13/0) \\
PTS & 14--0 (7/0) & 12--2 (10/0) & 14--0 (13/0) & 14--0 (12/0) \\
Dirichlet, ODIR & 12--2 (7/0) & 6--8 (6/4) & 12--2 (6/1) & 12--2 (10/0) \\
Linear mapper & 14--0 (10/0) & 9--5 (6/1) & 12--2 (9/0) & 14--0 (11/0) \\
MLP mapper & 14--0 (11/0) & 11--3 (6/0) & 14--0 (10/0) & 14--0 (14/0) \\
NAPHA & 14--0 (13/0) & 12--2 (10/0) & 14--0 (12/0) & 14--0 (14/0) \\
NAPHA, log-prob. & 14--0 (12/0) & 10--4 (6/0) & 12--2 (9/0) & 14--0 (11/0) \\
\citet{chen2025distributional}, replication & 1--13 (0/9) & 1--13 (0/8) & 1--13 (0/8) & 4--10 (1/8) \\
\hline
\end{tabular}
\hfill
\begin{tabular}{lcccccc}
\hline
\multicolumn{7}{l}{\textit{(b) Aggregated-label alignment, 11 ordinal benchmarks}} \\
HRS vs. & $\rho$ & Acc & F1 & MAE & Brier & ECE \\
\hline
G-Eval & 9--2 (4/0) & 11--0 (5/0) & 8--3 (5/1) & 11--0 (10/0) & 11--0 (11/0) & 11--0 (11/0) \\
LAGER & 11--0 (6/0) & 11--0 (6/0) & 10--1 (9/0) & 11--0 (8/0) & 10--1 (5/0) & 4--7 (2/1) \\
SAJA, regression & 11--0 (4/0) & 8--3 (6/0) & 10--1 (5/0) & 11--0 (6/0) & n/a & n/a \\
SAJA, classification & 10--1 (4/0) & 9--2 (5/0) & 9--2 (5/0) & 11--0 (6/0) & 4--7 (1/0) & 3--8 (0/4) \\
HypoEval & 10--0 (6/0) & 10--0 (8/0) & 9--1 (7/0) & 10--0 (10/0) & n/a & n/a \\
\citet{chen2025distributional}, replication & 1--10 (0/6) & 3--8 (1/3) & 2--9 (0/3) & 2--9 (0/3) & 3--8 (0/5) & 3--8 (0/1) \\
\hline
\end{tabular}
\caption{Seed-42 wins--losses of \method over benchmarks against each baseline, with the number of significant wins / significant losses in parentheses (Qwen2.5-7B). The \citet{chen2025distributional} row is our replication of their distributional LoRA fine-tuning. HypoEval does not apply to the pairwise HelpSteer3 task; scalar-only methods have no Brier score or ECE.}
\label{tab:wl}
\end{table*}

These complementary diagnostics do not imply universal calibration
superiority. In this seed-42 scope, \method loses to Dirichlet calibration
on 8 of 14 KL comparisons and to SAJA classification on 8 of 11 ECE
comparisons. Our primary distributional endpoint is JSD; KL, ECE, and the
other columns expose distinct trade-offs rather than interchangeable
confirmations of the same claim.

\begin{table*}[h]
\centering\scriptsize\setlength{\tabcolsep}{4pt}
\begin{tabular}{@{}l*{11}{c}c@{}}
\toprule
 & \multicolumn{4}{c}{\fontsize{5}{6}\selectfont SummEval} & \multicolumn{4}{c}{\fontsize{5}{6}\selectfont Topical-Chat} &  &  &  & \\
\cmidrule(lr){2-5}\cmidrule(lr){6-9}
Method & Coh & Con & Flu & Rel & Ovr & Nat & Ctx & Eng & FLASK & HS2 & HS3 & Avg. \\
\midrule
Raw judge & 1.070$^{\dagger}$ & 1.156$^{\dagger}$ & 1.705$^{\dagger}$ & .885$^{\dagger}$ & .844 & .666$^{\dagger}$ & .612$^{\dagger}$ & .489$^{\dagger}$ & .873$^{\dagger}$ & .866$^{\dagger}$ & 1.533$^{\dagger}$ & \multicolumn{1}{|c}{.973} \\
G-Eval~\citep{geval} & .949$^{\dagger}$ & .902$^{\dagger}$ & 2.116$^{\dagger}$ & .906$^{\dagger}$ & 1.388$^{\dagger}$ & .525 & .616$^{\dagger}$ & .406$^{\dagger}$ & .688$^{\dagger}$ & .859$^{\dagger}$ & 1.305$^{\dagger}$ & \multicolumn{1}{|c}{.969} \\
LAGER~\citep{lager} & .783$^{\dagger}$ & .833$^{\dagger}$ & .883$^{\dagger}$ & .564 & .867 & .514 & .511$^{\dagger}$ & .441$^{\dagger}$ & .675$^{\dagger}$ & .824$^{\dagger}$ & 1.292$^{\dagger}$ & \multicolumn{1}{|c}{.744} \\
SAJA, regression~\citep{saja} & .682$^{\dagger}$ & .379$^{\dagger}$ & .380$^{\dagger}$ & .533 & .794 & .462 & .463 & .469$^{\dagger}$ & .644 & .797$^{\dagger}$ & 1.178$^{\dagger}$ & \multicolumn{1}{|c}{.616} \\
SAJA, classification & .713$^{\dagger}$ & .366$^{\dagger}$ & .378$^{\dagger}$ & .526 & .830 & .444 & .464 & .485$^{\dagger}$ & .629 & .791$^{\dagger}$ & 1.180$^{\dagger}$ & \multicolumn{1}{|c}{.619} \\
HypoEval~\citep{hypoeval} & .856$^{\dagger}$ & 1.701$^{\dagger}$ & 2.264$^{\dagger}$ & .605$^{\dagger}$ & 1.177$^{\dagger}$ & .648$^{\dagger}$ & .729$^{\dagger}$ & .581$^{\dagger}$ & .894$^{\dagger}$ & .837$^{\dagger}$ & -- & \multicolumn{1}{|c}{1.029} \\
\midrule
\textbf{HRS (ours)} & \textbf{\underline{.566}} & \underline{.253} & \underline{.300} & \textbf{\underline{.493}} & \underline{.776} & \underline{.439} & \underline{.424} & \underline{.347} & \underline{.584} & \underline{.706} & \underline{1.018} & \multicolumn{1}{|c}{\underline{.537}} \\
\citet{chen2025distributional}, replication & .578 & \textbf{.243} & \textbf{.279} & .499 & \textbf{.666}$^{\dagger}$ & \textbf{.385} & \textbf{.414} & \textbf{.337} & \textbf{.541} & \textbf{.664}$^{\dagger}$ & \textbf{.867}$^{\dagger}$ & \multicolumn{1}{|c}{\textbf{.498}} \\
\bottomrule
\end{tabular}
\caption{Aggregated scores: MAE between the expected score and the mean human rating on the 11 ordinal benchmarks (Qwen2.5-7B; lower is better). Marks as in Table~\ref{tab:h2ft}.}
\label{tab:h2mae}
\end{table*}

\section{Cross-Judge Replication and Fine-Tuning}
\label{app:replication_finetuning}

\subsection{Comparison with Weight Fine-Tuning}
\label{app:finetune}
LoRA adapters~\citep{lora} are trained in every transformer block. The 14-benchmark study uses our re-implementation of the distribution-alignment fine-tuning of \citet{chen2025distributional} (objective and recipe in Appendix~\ref{app:expbaselines}; no code is released, and the adapter shape is our choice). Under the primary JSD evaluation, the replication of \citet{chen2025distributional} has the lower score on 13 of 14 primary-judge benchmarks, averaging $0.338$ against $0.352$ for HRS across three seeds. HRS therefore provides a parameter-efficient alternative to backbone fine-tuning rather than stronger distributional alignment.

Table~\ref{tab:time14} reports the corresponding fitting and cache-construction trade-off across all 14 benchmarks. On HelpSteer2, the replication trains 10.1M parameters through all 28 blocks in 210--240 minutes and obtains annotation-level RPS $0.616$. HRS updates at most 115k parameters, about $88\times$ fewer, while preserving the pretrained backbone; its five frozen upper blocks require a 52-minute, 120-GB prefix cache and 17 minutes of fitting. The resulting 69 minutes is a fitting-and-cache subtotal, not the full training cost: constructing the oracle corrections used by the reported initialisation adds about two hours on HelpSteer2 when they are not already available, for roughly 189 minutes end to end. The fitting-time advantage is also setting-dependent: in exact mode, which SummEval and ChaosNLI use, HRS recomputes the judge forward pass every epoch and takes 39--70 minutes per SummEval criterion and 238 minutes on ChaosNLI, against 56--83 and 111 minutes for the replication of \citet{chen2025distributional}. Across the suite, 14.2 GPU-hours is likewise a fitting-and-cache subtotal that excludes oracle construction, whereas the 19.8 GPU-hours for the replication includes its training. The parameter-count comparison is therefore the clean efficiency result; the wall-clock comparison is not end-to-end matched. The single-layer label-dependent oracle reaches 0.407, demonstrating corrective capacity rather than deployable
prediction. At inference, LoRA requires one adapted
judge forward pass, whereas HRS additionally re-evaluates
the frozen downstream layers after computing the original
rating logits. HRS therefore incurs additional inference
computation while retaining removable, criterion-specific
interventions without modifying backbone weights.

\subsection{Cross-Judge and Cross-Seed Results}
\label{app:llama}
Table~\ref{tab:seeds14} reports variability across the three Qwen2.5-7B runs used for the primary results.
\begin{table*}[h]
\centering\scriptsize\setlength{\tabcolsep}{3pt}
\resizebox{0.49\textwidth}{!}{%
\begin{tabular}{lccc}
\hline
Benchmark & Dirichlet & \method & \citet{chen2025distributional} \\
\hline
SummEval coherence & .473 $\pm$ .012 & .414 $\pm$ .018 & \textbf{.407 $\pm$ .019} \\
SummEval consistency & .209 $\pm$ .014 & \textbf{.159 $\pm$ .033} & .179 $\pm$ .030 \\
SummEval fluency & .288 $\pm$ .003 & .241 $\pm$ .008 & \textbf{.229 $\pm$ .029} \\
SummEval relevance & .401 $\pm$ .016 & .383 $\pm$ .019 & \textbf{.379 $\pm$ .019} \\
Topical-Chat overall & .473 $\pm$ .002 & .440 $\pm$ .014 & \textbf{.434 $\pm$ .002} \\
Topical-Chat natural & .376 $\pm$ .011 & .379 $\pm$ .010 & \textbf{.341 $\pm$ .006} \\
Topical-Chat context & .352 $\pm$ .005 & .349 $\pm$ .014 & \textbf{.338 $\pm$ .026} \\
\hline
\end{tabular}}
\hfill
\resizebox{0.49\textwidth}{!}{%
\begin{tabular}{lccc}
\hline
Benchmark & Dirichlet & \method & \citet{chen2025distributional} \\
\hline
Topical-Chat engaging & .345 $\pm$ .009 & .321 $\pm$ .005 & \textbf{.292 $\pm$ .006} \\
FLASK & .425 $\pm$ .005 & .410 $\pm$ .004 & \textbf{.397 $\pm$ .010} \\
HelpSteer2 helpfulness & .448 $\pm$ .001 & .432 $\pm$ .003 & \textbf{.417 $\pm$ .003} \\
HelpSteer3 code & .576 $\pm$ .003 & .546 $\pm$ .004 & \textbf{.519 $\pm$ .003} \\
ChaosNLI & .214 $\pm$ .000 & .198 $\pm$ .002 & \textbf{.184 $\pm$ .008} \\
DynaSent R2 & .324 $\pm$ .002 & .302 $\pm$ .008 & \textbf{.269 $\pm$ .002} \\
Scruples Anecdotes & .375 $\pm$ .000 & .348 $\pm$ .002 & \textbf{.343 $\pm$ .008} \\
\hline
Mean & .377 & .352 & .338 \\
\hline
\end{tabular}}
\caption{Three-seed variability in the Qwen2.5-7B study: mean $\pm$ standard deviation of test JSD over seeds 42, 13, and 7 (the \method column is the primary row of Table~\ref{tab:h1}). Splits are re-drawn by seed except for ChaosNLI and Scruples. The \citet{chen2025distributional} column is our replication of their distributional LoRA fine-tuning. \textbf{Bold}: best method.}
\label{tab:seeds14}
\end{table*}

\section{Generalisation and Data Requirements}
\label{app:generalisation}

The following experiments are preliminary: transfer is restricted to seven
compatible five-category ordinal benchmarks, and the label-budget study to
three settings. They diagnose dependence on in-domain supervision rather
than establish broad cross-domain or annotation-efficient adaptation.

\paragraph{Cross-benchmark transfer.}
\label{app:xfer}
The seven ordinal benchmarks with five categories permit direct transfer while retaining every fitted parameter, including the HRS layer and scale. Table~\ref{tab:generalisation} shows that transfer improves on the raw judge but remains well behind in-domain training: Dirichlet calibration transfers best ($0.732$ mean RPS), while HRS reaches $0.850$ versus $0.473$ in-domain. Transfer is criterion-dependent; SummEval fluency to consistency reaches $0.201$ (in-domain $0.199$), whereas modules trained on HelpSteer2 or Topical-Chat degrade several SummEval criteria.

\paragraph{Label requirements.}
\label{app:budget}
With 100 training items, Dirichlet calibration is within $0.011$ of its full-data score on all three budget benchmarks. HRS is strongest on SummEval coherence ($0.576$), but at roughly 1,000 DynaSent items the replication of \citet{chen2025distributional} reaches $0.451$ versus $0.482$ for HRS. Output calibration is therefore competitive at small budgets, while learned representations benefit more from additional labels. Validation annotations are additional to the stated training budgets.

\begin{table*}[h]
\centering\small
\begin{tabular}{llccccc}
\hline
Setting & Training items & Raw & TS-Soft & Dirichlet & HRS & \citet{chen2025distributional} \\
\hline
Cross-benchmark transfer, mean & another benchmark & 1.096 & .825 & \textbf{.732} & .850 & .786 \\
In-domain, same seven benchmarks & full & 1.096 & -- & .489 & .473 & \textbf{.450} \\
FLASK & 100 & .960 & .758 & \textbf{.557} & .566 & .575 \\
SummEval coherence & 100 & 1.095 & .720 & .611 & \textbf{.576} & .602 \\
DynaSent R2 & 100 & .728 & .536 & \textbf{.501} & .505 & .510 \\
DynaSent R2 & 1,000 & .728 & .527 & .491 & .482 & \textbf{.451} \\
\hline
\end{tabular}
\caption{Compact transfer and label-budget results (Qwen2.5-7B; annotation-level proper score, lower is better). Transfer averages each target over six source benchmarks. Budget rows refit every method on the same prefix of the training data while retaining the full validation split. The \citet{chen2025distributional} column is our replication of their distributional LoRA fine-tuning.}
\label{tab:generalisation}
\end{table*}
\clearpage
\section{Oracle Correctability and Reproducibility}
\label{app:oracle_reproducibility}

\subsection{Oracle Solver and Low-Rank Diagnostics}
\label{app:oracle}
\paragraph{Computation.} Algorithm~\ref{alg:oracle} gives the implemented three-step damped logit-correction solver. Vanilla Jacobians are cached in float16; re-linearised Jacobians are computed in bfloat16, retained in float32, and the $K\times K$ systems are solved in float32 (float64 for the first step). Relative damping is fixed at $0.01$; later steps use batches of at most eight items and 6,000 tokens. The five candidate depths are $\{0.2,0.4,0.6,0.8,0.95\}L$, selected on validation. Oracle corrections use the evaluated item's ratings and are diagnostic, not deployable. The residual is centred by $C=I-\mathbf 1\mathbf 1^\top/K$, but the implementation uses the uncentred Jacobian $G=\partial g_l/\partial h$ in the update below. It is therefore not exactly the damped Gauss--Newton update for the centred least-squares objective, which would use $\Delta\delta=(CG)^\top[(CG)(CG)^\top+\lambda I]^{-1}r$ (and scale the relative damping from $CG$). All reported oracle and SVD results use the implemented form shown here.
\begin{algorithm}[h]
\caption{Implemented oracle residual $\delta^\ast(x)$ of one item (\S\ref{sec:oracle}).}
\label{alg:oracle}
\small
\begin{algorithmic}[1]
\REQUIRE item $x$ with smoothed histogram $\widetilde P_H$; block $l$; $\varepsilon=10^{-4}$; relative damping $0.01$
\STATE $z_H\leftarrow\log(\widetilde P_H+\varepsilon\mathbf 1)$; $\delta\leftarrow0$
\FOR{$n=0,1,2$}
\STATE $r\leftarrow C\big(z_H-g_l(h_l(x)+\delta;x)\big)$ \hfill (for $n=0$: the cached vanilla logits $z(x)$)
\STATE $G\leftarrow\partial g_l/\partial h$ at $h_l(x)+\delta$ by $K$ backward passes \hfill (for $n=0$: the cached $G(0)$)
\STATE $\lambda_n\leftarrow0.01\cdot\overline{\mathrm{tr}\,GG^\top}/K$ \hfill (mean over the split for $n=0$, over the mini-batch otherwise)
\STATE $\delta\leftarrow\delta+G^\top\big(GG^\top+\lambda_nI_K\big)^{-1}r$
\ENDFOR
\STATE log $\|r\|_2$ at the final $\delta$ (one more forward pass); \textbf{return} $\delta^\ast=\delta$
\end{algorithmic}
\end{algorithm}
\paragraph{Correctability and low-rank structure.} Table~\ref{tab:oraclesummary} consolidates the test oracle, rank-8 projection, residual spectrum, and validation depth sweep. Full corrections reduce mean test JSD from $0.565$ to $0.179$ for Qwen (mean of seeds 42, 13, and 7; per seed $0.563\to0.186$, $0.573\to0.167$, and $0.558\to0.183$) and from $0.490$ to $0.184$ for Llama (seed 42). Rank eight retains $84\%$ and $99\%$ of these improvements, respectively. The result is weaker for Qwen: its projected JSD is $0.241$, and rank eight captures a median $94\%$ of residual energy (seed 42), versus $0.185$ and at least $98.8\%$ for Llama. Validation performance is essentially flat from $0.6L$ onward. These are label-dependent corrections; learning their coefficients remains the predictive problem. Moreover, the benchmarks have only 3--6 rating logits. Rank eight is a shared subspace in the much larger hidden-state space, but oracle correctability in this small output space is motivation rather than evidence that the oracle initialisation causes predictive gains.

\begin{table*}[t]
\centering\small
\resizebox{\textwidth}{!}{%
\begin{tabular}{lcccclccccc}
\hline
 & \multicolumn{3}{c}{Test JSD, 14 benchmarks} & & \multicolumn{5}{c}{Validation oracle JSD by relative depth} \\
Judge & Raw & Full oracle & Rank-8 oracle & Rank-8 energy & $0.2L$ & $0.4L$ & $0.6L$ & $0.8L$ & $0.95L$ \\
\hline
Qwen2.5-7B & .565 & .179 & .241 & 94\% median & .242 & .201 & .194 & .193 & .203 \\
Llama-3.1-8B & .490 & .184 & .185 & 98.8--99.7\% & .196 & .186 & .186 & .186 & .186 \\
\hline
\end{tabular}}
\caption{Compact oracle diagnostics. Test columns average all 14 benchmarks; for Qwen2.5-7B they are the mean of seeds 42, 13, and 7, for Llama-3.1-8B seed 42. The energy and depth columns use seed 42. Depth columns are validation means over the 12 Qwen and 13 Llama benchmarks with recorded JSD at every candidate depth. Rank-8 energy summarises the training-residual spectrum. Full and projected oracles use test-item ratings and are not deployable predictors.}
\label{tab:oraclesummary}
\label{tab:oracledepthcomplete}
\label{tab:oraclespectrumcomplete}
\end{table*}

\subsection{Predictive Component Ablations}
\label{app:component_ablations}
Table~\ref{tab:components_short} isolates the components most directly tied
to the oracle motivation. At matched rank eight, random orthonormal
initialisation is effectively tied with oracle-SVD initialisation: it is
$0.001$ worse under annotation-level RPS and $0.004$ better under JSD. The
rank sweep is similarly flat from rank two through sixteen. A separate
matched rank-4 run that changes only whether $V$ is updated obtains RPS
$0.668$ with the oracle basis frozen and $0.646$ with it trainable, so
optimising the basis matters more in that control than its initialisation.

\begin{table}[h]
\centering\small
\resizebox{\columnwidth}{!}{%
\begin{tabular}{lrrr}
\toprule
Variant & Rank & JSD$\downarrow$ & Ann. RPS$\downarrow$ \\
\midrule
Reported \method, oracle-SVD init. & 16 & .431 & .641 \\
Oracle-SVD init., matched control & 8 & .433 & .644 \\
Random orthonormal init. & 8 & \textbf{.429} & .645 \\
Router without original logits & 8 & .435 & \textbf{.642} \\
Oracle-SVD init. & 2 & .433 & .646 \\
Oracle-SVD init. & 4 & .439 & .649 \\
\bottomrule
\end{tabular}}
\caption{Targeted component controls on HelpSteer2 with Qwen2.5-7B, seed 42. All rows train $V$ end to end and use the same data and optimisation protocol; the matched rank-8 SVD and random rows both initialise the router randomly. The reported rank-16 row is included as a reference, not as part of the matched initialisation comparison.}
\label{tab:components_short}
\end{table}

These single-benchmark controls do not establish that the oracle SVD is
necessary for learned \method; in this setting it is not. We therefore use
the oracle analysis to motivate the intervention geometry and demonstrate
corrective capacity, not as a causal explanation of predictive performance.
All headline runs nevertheless use oracle-SVD initialisation, so the cost
remains part of the reported pipeline. Removing the original logits also
does not hurt here, showing that the hidden representation is sufficient in
this control; an exactly capacity-matched logit-only router remains untested.

\subsection{Training Details of \method}
\label{app:training}
Algorithm~\ref{alg:hrs} summarises the training of \method for one judge and one benchmark. Its item loss $\bar S(x)$ is the mean annotation-level RPS for ordinal tasks and $S_{\mathrm{CE}}(\PHh(\cdot\mid x),\widetilde P_H(\cdot\mid x))$ for nominal tasks, as specified in \S\ref{sec:training}.
\begin{algorithm}[h]
\caption{Training of \method.}
\label{alg:hrs}
\small
\begin{algorithmic}[1]
\REQUIRE frozen judge $J$; training and validation items with individual ratings; rank $r$; label type
\STATE cache $h_l(x)$ and $z(x)$ of the vanilla judge at the candidate depths
\STATE compute the oracle corrections $\delta^\ast(x)$ of the training items; select $l$ on validation
\STATE $V\leftarrow$ the $r$ leading right-singular vectors of $\{\delta^\ast(x)\}$; draw $W_a, b_a$ at random
\STATE (long inputs) cache the prefix keys and values of the blocks above $l$
\REPEAT
\FOR{each mini-batch $B$ of training items}
\STATE $a\leftarrow W_a[\mathrm{LN}(h_l);\tilde z]+b_a$; $\;h_l'\leftarrow h_l+Va$
\STATE $\PHh\leftarrow\mathrm{softmax}\big(g_l(h_l')\big)$ through the frozen blocks above $l$ (single token, constant prefix)
\STATE update $(V,W_a,b_a)$ by AdamW on $\frac{1}{|B|}\sum_{x\in B}\big[\bar S(x)+\lambda_a\|a(x)\|^2\big]$
\ENDFOR
\STATE score the edited judge on the validation items; keep the best epoch
\UNTIL{no improvement for \emph{patience} epochs, or the epoch cap is reached}
\STATE select $\alpha$ on the validation split; \textbf{return} $(V,W_a,b_a,l,\alpha)$
\end{algorithmic}
\end{algorithm}
\paragraph{Objective and optimisation.} Averaging RPS over an item's ratings equals squared distance between the predicted and empirical cumulative distributions plus a target-dependent constant; thus Eq.~\eqref{eq:loss} uses the raw annotations without requiring a smoothed training target. Nominal tasks use cross-entropy to an empirical-Bayes-smoothed histogram. We train $V$, the linear router, and its bias with AdamW (learning rate $5\times10^{-4}$, batch 32 for cached long inputs, $\lambda_a=10^{-4}$), validate after each epoch, and early-stop according to Table~\ref{tab:hrsettings}. Rank and $\alpha\in\{0.5,0.75,1,1.25,1.5\}$ are selected on validation.

The router is initialised at random; the basis remains trainable and is not constrained to stay orthogonal.
\subsection{Implementation Details}
\label{app:details}
\paragraph{Inputs and targets.} The zero-shot prompt names the criterion and its $K$ verbalised scale points and ends with ``Return only one rating token:''; the distribution is the softmax over those single-token verbalizers. Inputs are truncated to 1,536 tokens (2,304 for code pairs; 512 for comments/posts). Histogram smoothing shrinks toward the training marginal with strength selected by validation leave-one-out RPS for ordinal labels and Brier score for nominal labels; evaluation always uses the raw annotations.

\paragraph{Prompt template.}\label{app:prompt_template}
For each criterion, the raw judge, \method, output calibrators, LAGER, and distributional LoRA comparisons share the same zero-shot prompt across judge backbones. G-Eval, SAJA, and HypoEval use the method-specific prompts described in Appendix~\ref{app:expbaselines}. The shared response form is:
\begin{quote}
\small\ttfamily
You are evaluating the quality of an assistant response.\par
\medskip
Criterion: [criterion]\par
\medskip
Rate the response on a [K]-point scale:\par
A = [lowest scale description]\par
\ldots\par
[Kth token] = [highest scale description]\par
\medskip
User prompt:\par
[user prompt]\par
\medskip
Assistant response:\par
[response]\par
\medskip
Return only one rating token:
\end{quote}
For summaries, dialogues, pairwise comparisons, and nominal classification, only the task-specific field names and the final word \emph{rating}/\emph{label} change; the options remain single-token letters.

\paragraph{Model selection and baselines.} The candidate depths, judge-specific layer-selection rules, ranks, and scale grid are specified in Appendix~\ref{app:expbaselines}; selected seed-42 settings appear in Table~\ref{tab:hrsettings}. Baseline optimisers, prompts, and handling of unparseable outputs are method-specific, as described in Appendix~\ref{app:expbaselines}.

\paragraph{Compute.} Experiments use Python 3.10.12, PyTorch 2.6.0, Transformers 4.51.1, bfloat16 judges, and RTX 3090 GPUs; Qwen2.5-14B uses two GPUs. A partial prefix store is 120 GB and takes 52 minutes for HelpSteer2; after this preprocessing, HRS fitting takes about 17 minutes. These figures exclude the approximately two hours needed to construct the HelpSteer2 oracle initialisation when it is not already available. Short-input datasets use exact forward computation. Table~\ref{tab:time14} reports fitting and cache-construction times, not end-to-end runtime including oracle construction.
\subsection{Selected Settings of \method}
\label{app:hrsettings}
All 80 in-design judge--seed--benchmark cells have primary HRS and output-calibration results. Qwen2.5-14B intentionally covers ten benchmarks; HelpSteer2, HelpSteer3, ChaosNLI, and Scruples are outside that configuration. A dash elsewhere denotes an inapplicable or unrun secondary comparison, not a pending primary result. Table~\ref{tab:hrsettings} records validation-selected layer, rank, scale, loss, stopping rule, and cache use for every seed-42 configuration; seeds 13 and 7 select their own layers and scales.
\begin{table*}[h]
\centering\scriptsize\setlength{\tabcolsep}{3.5pt}
\resizebox{\textwidth}{!}{%
\begin{tabular}{lcccccccccccclcc}
\hline
 & \multicolumn{3}{c}{Qwen2.5-7B} & \multicolumn{3}{c}{Llama-3.1-8B} & \multicolumn{3}{c}{Gemma-2-9B} & \multicolumn{3}{c}{Qwen2.5-14B} & & & \\
Benchmark & $l$ & $r$ & $\alpha$ & $l$ & $r$ & $\alpha$ & $l$ & $r$ & $\alpha$ & $l$ & $r$ & $\alpha$ & Loss & Epochs / patience & Store \\
\hline
Coh & 22 & 8 & 1 & 30 & 8 & 1.25 & 40 & 8 & 1 & 45 & 8 & 1 & CDF & 40 / 8 & n/a \\
Con & 16 & 8 & 0.75 & 30 & 8 & 1.5 & 40 & 8 & 1 & 45 & 8 & 1 & CDF & 40 / 8 & n/a \\
Flu & 22 & 8 & 1 & 30 & 8 & 1 & 40 & 8 & 1 & 45 & 8 & 1 & CDF & 40 / 8 & n/a \\
Rel & 22 & 8 & 1 & 30 & 8 & 1 & 40 & 8 & 1 & 45 & 8 & 1 & CDF & 40 / 8 & n/a \\
Ovr & 22 & 8 & 1 & 30 & 8 & 0.75 & 40 & 8 & 1 & 45 & 8 & 1 & CDF & 60 / 10 & n/a \\
Nat & 16 & 8 & 1 & 30 & 8 & 1.25 & 40 & 8 & 1 & 45 & 8 & 1 & CDF & 60 / 10 & n/a \\
Ctx & 22 & 8 & 1 & 30 & 8 & 1 & 40 & 8 & 1 & 45 & 8 & 1.25 & CDF & 60 / 10 & n/a \\
Eng & 11 & 8 & 1.25 & 30 & 8 & 1 & 40 & 8 & 1 & 45 & 8 & 1 & CDF & 60 / 10 & n/a \\
FLASK & 22 & 8 & 1 & 30 & 8 & 1.5 & 40 & 8 & 1 & 45 & 8 & 1.25 & CDF & 40 / 8 & n/a \\
HS2 & 22 & 16 & 1 & 30 & 16 & 1.25 & 40 & 16 & 1 & n/a & n/a & n/a & CDF & 30 / 5 & \checkmark \\
HS3 & 22 & 8 & 0.75 & 30 & 8 & 0.5 & 40 & 8 & 1 & n/a & n/a & n/a & CDF & 30 / 5 & \checkmark \\
CNLI & 22 & 8 & 1 & 30 & 8 & 1.5 & 40 & 8 & 1.25 & n/a & n/a & n/a & CE & 20 / 6 & n/a \\
DynaS & 22 & 8 & 0.75 & 30 & 8 & 0.75 & 40 & 8 & 1 & 45 & 8 & 0.75 & CE & 20 / 6 & n/a \\
Scrup & 22 & 8 & 1 & 30 & 8 & 1 & 40 & 8 & 1 & n/a & n/a & n/a & CE & 30 / 10 & \checkmark \\
\hline
\end{tabular}}
\caption{Settings of the primary \method row (seed 42) per judge: steering block $l$ (of 28, 32, 42 and 48 blocks), rank $r$ and scale $\alpha$, all selected on validation; the training loss (CDF: annotation RPS; CE: cross-entropy to the smoothed histogram), the cap on epochs and the patience, and whether training used a partial key--value store. n/a: outside the design of the run.}
\label{tab:hrsettings}
\label{tab:coverage}
\end{table*}

\subsection{Compute and Runtime}
\label{app:cost}
Table~\ref{tab:time14} compares HRS fitting plus cache construction with our distributional LoRA replication on all 14 benchmarks. On that restricted accounting, HRS is faster on 11 benchmarks and sums to 14.2 GPU-hours, including three cache builds, versus 19.8 GPU-hours for LoRA. The slower HRS cases are Topical-Chat context, ChaosNLI, and DynaSent, where exact-mode training repeatedly evaluates the largest training sets. The 14.2-hour subtotal excludes construction of the oracle corrections required by every reported primary HRS run; this adds about two hours on HelpSteer2 when computed afresh, and corresponding oracle costs are not aggregated for the full suite. We therefore do not interpret the two totals as an end-to-end speedup. At inference, HRS first computes the original judge
representations and rating logits, then propagates
the steered final-token representation through the frozen
downstream layers. Unlike LoRA, whose adapters can be
merged into the backbone for a single adapted forward
pass, HRS requires this additional suffix computation.
Its overhead depends on the intervention depth and
implementation, and is not included in the training-time
comparison.

\begin{table}[h]
\centering\scriptsize\setlength{\tabcolsep}{3pt}
\resizebox{\columnwidth}{!}{%
\begin{tabular}{llrrrrrr}
\hline
Benchmark & HRS mode & Epochs & HRS fit & Store build (GB) & Fit + store & \citet{chen2025distributional} & HRS / \citet{chen2025distributional} \\
\hline
SummEval coherence & exact & 21 & 70 & -- & 70 & 76 & 0.92 \\
SummEval consistency & exact & 12 & 39 & -- & 39 & 56 & 0.70 \\
SummEval fluency & exact & 19 & 66 & -- & 66 & 83 & 0.79 \\
SummEval relevance & exact & 18 & 62 & -- & 62 & 65 & 0.96 \\
Topical-Chat overall & exact & 22 & 13 & -- & 13 & 15 & 0.86 \\
Topical-Chat natural & exact & 19 & 11 & -- & 11 & 15 & 0.74 \\
Topical-Chat context & exact & 27 & 17 & -- & 17 & 12 & 1.49 \\
Topical-Chat engaging & exact & 22 & 13 & -- & 13 & 15 & 0.88 \\
FLASK & exact & 11 & 53 & -- & 53 & 80 & 0.66 \\
HelpSteer2 helpfulness & cached & 18 & 17 & 52 (120) & 69 & 223 & 0.31 \\
HelpSteer3 code & cached & 12 & 7.8 & 46 (108) & 54 & 233 & 0.23 \\
ChaosNLI & exact & 18 & 238 & -- & 238 & 111 & 2.14 \\
DynaSent R2 & exact & 16 & 107 & -- & 107 & 66 & 1.62 \\
Scruples Anecdotes & cached & 27 & 13 & 27 (64) & 40 & 136 & 0.29 \\
\hline
Subtotal (excl. oracle construction) & & & & & 14.2 h & 19.8 h & 0.72 \\
\hline
\end{tabular}}
\caption{Training time of the primary \method row against our
replication of the distributional LoRA fine-tuning of
\citet{chen2025distributional} on all 14 benchmarks (Qwen2.5-7B, seed
42, one RTX 3090; minutes unless stated). HRS fit sums the
per-epoch times of the promoted run; the store build is the
partial key--value cache of the three cached benchmarks (train,
validation, and test splits; size in GB); Fit + store adds the
two. \citet{chen2025distributional}: recorded wall-clock training time.
The reported HRS runs require oracle-SVD initialisation, whose construction
is excluded here (about two additional hours on HelpSteer2 when computed
afresh; Appendix~\ref{app:cost}).}
\label{tab:time14}
\end{table}

\section{Additional Diagnostics and Downstream Utility}
\label{app:additional_diagnostics}

\subsection{Preservation Reliability}
\label{app:harm}
On HelpSteer2, all trained methods harm some initially accurate predictions: among the quarter of items with the lowest vanilla-judge RPS, harm rates are $0.71$--$0.84$. A post-hoc gate that mixes vanilla and HRS distributions traces an RPS--harm frontier from $(0.645,0.67)$ to $(0.720,0.49)$; scaling the intervention gives a similar trade-off. A label-dependent oracle gate reaches $(0.602,0.00)$, so preservation requires a second per-item prediction problem. This limitation qualifies both HRS and fine-tuning gains.

\subsection{Downstream Statistical Efficiency}
\label{app:extra}
\begin{table*}[!t]
\centering\small
\begin{tabular}{lrrcc}
\toprule
 & \multicolumn{2}{c}{ESS gain versus human-only ($\%$)}
 & \multicolumn{2}{c}{Coverage of 95\% intervals} \\
Judge probabilities & HS2 & HS3 code & HS2 & HS3 code \\
\midrule
Raw judge & +2 & $-40$ & .950 & .946 \\
Dirichlet calibration & +21 & +26 & .956 & .953 \\
NAPHA & +16 & +9 & .954 & .953 \\
\textbf{HRS} & +38 & +39 & .951 & .952 \\
\bottomrule
\end{tabular}
\caption{Population estimation with uniform-sampling ASI and
a 10\% expected evaluation-item annotation budget, using
Qwen2.5-7B (seed 42). Gains are ratios of Monte Carlo
variances, aggregated across five calibration/evaluation
partitions. Calibration
and model-training labels are outside the evaluation budget.}
\label{tab:population_efficiency}
\end{table*}

\paragraph{Target of estimation.}
We replicate active statistical inference~\citep{fsa_asi} on
HelpSteer2 and HelpSteer3 code with Qwen2.5-7B (seed 42).
For an evaluation pool of $N$ items, let $h_i$ be the fraction
of item $i$'s human ratings that are at least 3, and let
$f_i=\sum_{k\ge3}\widehat P_H(k\mid x_i)$ be the corresponding
judge probability. The target is
$\theta=N^{-1}\sum_i h_i$, averaging items equally.
On HelpSteer2 this is the rate of mostly or extremely helpful
judgments; on HelpSteer3 code, preference for response 2.
The target is a population statistic of human judgments,
including their disagreement, rather than a majority-label
accuracy or an average of JSD values.

\paragraph{Judge-assisted estimator.}
An item receives human annotation when
$\xi_i\sim\mathrm{Bernoulli}(\pi_i)$. With a coefficient
$\lambda$ fixed on separate calibration data, the estimator is
\begin{equation}
\widehat\theta=\frac{1}{N}\sum_{i=1}^{N}
\left[\lambda f_i+\frac{\xi_i}{\pi_i}
\big(h_i-\lambda f_i\big)\right].
\label{eq:population_estimator}
\end{equation}
The first term uses predictions for every item; the second
estimates their residual error from the sampled human ratings.
Inverse-probability weighting corrects for unequal sampling.
For fixed $\lambda$, positive $\pi_i$, and independent
Bernoulli draws, the estimator is unbiased over annotation
sampling for the finite-pool target, with variance
\begin{equation}
\operatorname{Var}(\widehat\theta)=\frac{1}{N^2}
\sum_i\big(h_i-\lambda f_i\big)^2
\frac{1-\pi_i}{\pi_i}.
\label{eq:population_variance}
\end{equation}
More reliable judge probabilities can therefore reduce
estimation variance through smaller residuals. Active policies
can additionally allocate more labels to items predicted to
have large residuals. A low JSD alone does not guarantee
either benefit, so we measure downstream variance directly.

\paragraph{Simulation and annotation budget.}
Using the released annotations, we form five prompt-level
40/60 calibration/evaluation partitions of each test set.
All calibration items are labelled; this split determines
$\lambda$ and any sampling-policy parameters. For each
partition and estimator, we simulate 1,000 independent
annotation draws on the remaining evaluation pool. Uniform
sampling sets $\pi_i=0.1$, so the expected evaluation budget
is 10\% of that pool. The calibration annotations and
the earlier model-training annotations are additional to
this budget. This experiment evaluates sampling efficiency
conditional on those resources, not an end-to-end 10\%
annotation requirement. Confidence intervals use the
plug-in sampling-variance estimate and a normal approximation.

\paragraph{Effective sample size and results.}
The human-only reference averages ratings from uniformly
sampled evaluation items. We report the variance-based gain
\begin{equation}
G_{\mathrm{ESS}}=
\frac{\operatorname{Var}(\widehat\theta_{\mathrm{human}})}
{\operatorname{Var}(\widehat\theta)}-1.
\label{eq:population_ess}
\end{equation}
A gain of 39\% corresponds to 1.39 times the reference
effective sample size, or about 28\% lower variance at
the same evaluation-label budget. Table~\ref{tab:population_efficiency}
reports uniform-sampling ASI, isolating the choice of judge
from the annotation policy. HRS improves over the raw judge and
output calibration on both tasks. This downstream result is limited to two related HelpSteer
tasks and should not be read as benchmark-general evidence of annotation
efficiency.
The failure-space policy changes ESS by $-1\%$ on HelpSteer2
and $+1\%$ on HelpSteer3 code relative to uniform ASI with HRS. Empirical coverage is the
fraction of simulated intervals containing the full-pool
human target and remains close to the nominal 95\% level.

\end{document}